\documentclass{article}

\PassOptionsToPackage{numbers, compress}{natbib}

 \usepackage[preprint]{neurips_2026}

\usepackage[utf8]{inputenc} 
\usepackage[T1]{fontenc}    
\usepackage{url}            
\usepackage{graphicx}       
\usepackage{subcaption}     
\usepackage{caption}        
\usepackage{booktabs}       
\usepackage{threeparttable}
\usepackage{tabularx}
\usepackage{amsmath}
\usepackage{amsfonts}       
\usepackage{amssymb}
\usepackage{mathtools}
\usepackage{amsthm}
\usepackage{longtable}
\usepackage{multirow}
\usepackage{enumitem}
\usepackage{needspace}
\usepackage{wrapfig}
\usepackage{bm}
\usepackage{nicefrac}       
\usepackage{microtype}      
\usepackage[table]{xcolor}  
\usepackage{hyperref}       
\usepackage[capitalize,noabbrev]{cleveref}

\def\ourMethod{{\textit{Robust Dreamer}}}

\newcommand{\secref}[1]{Sec.~\ref{#1}}
\newcommand{\figref}[1]{Fig.~\ref{#1}}
\newcommand{\tabref}[1]{Tab.~\ref{#1}}
\renewcommand{\eqref}[1]{Eq.~\ref{#1}}

\definecolor{lightgreen}{rgb}{0.,0.5,0.}
\definecolor{remark}{rgb}{1,.5,0}
\definecolor{citecolor}{rgb}{0,0.443,0.737}
\definecolor{linkcolor}{rgb}{0.956,0.298,0.235}
\definecolor{myblue}{RGB}{79,113,190}
\definecolor{cvprblue}{rgb}{0.21,0.49,0.74}
\definecolor{lightred}{rgb}{0.5,0.,0.}

\theoremstyle{plain}

\theoremstyle{definition}

\theoremstyle{remark}

\usepackage[disable,textsize=tiny]{todonotes}

\definecolor{citecolor}{rgb}{0,0.443,0.737}
\definecolor{linkcolor}{rgb}{0.956,0.298,0.235}

\hypersetup{
    colorlinks=true,
    citecolor=citecolor,
    filecolor=citecolor,
    linkcolor=linkcolor,
    urlcolor=magenta
}

\title{Robust Dreamer: Deviation-Aware Latent Gaussian Memory for Action-Controlled AR Video Generation}

\author{%
    \textbf{Hanlin Chen$^{1}$\!\!\quad Jiaxin Wei$^{2}$\!\!\quad Xibin Song$^{3}$\!\!\quad Yifu Wang$^{3}$} \\
    \textbf{Steve Wang$^{3}$\!\!\quad Hongdong Li$^{4}$\!\!\quad Pan Ji$^{3}$\!\!\quad Gim Hee Lee$^{1}$} \\
    $^{1}$\, School of Computing, National University of Singapore \quad
    $^{2}$\, Technische Universit\"at M\"unchen \\
    $^{3}$\, Vertex Lab \quad
    $^{4}$\, Australian National University \\
    \texttt{hanlin.chen@u.nus.edu} \\
}

\begin{document}

    \maketitle


\begin{center}
    \begin{minipage}{\textwidth} 
        \centering
        \includegraphics[width=\textwidth]{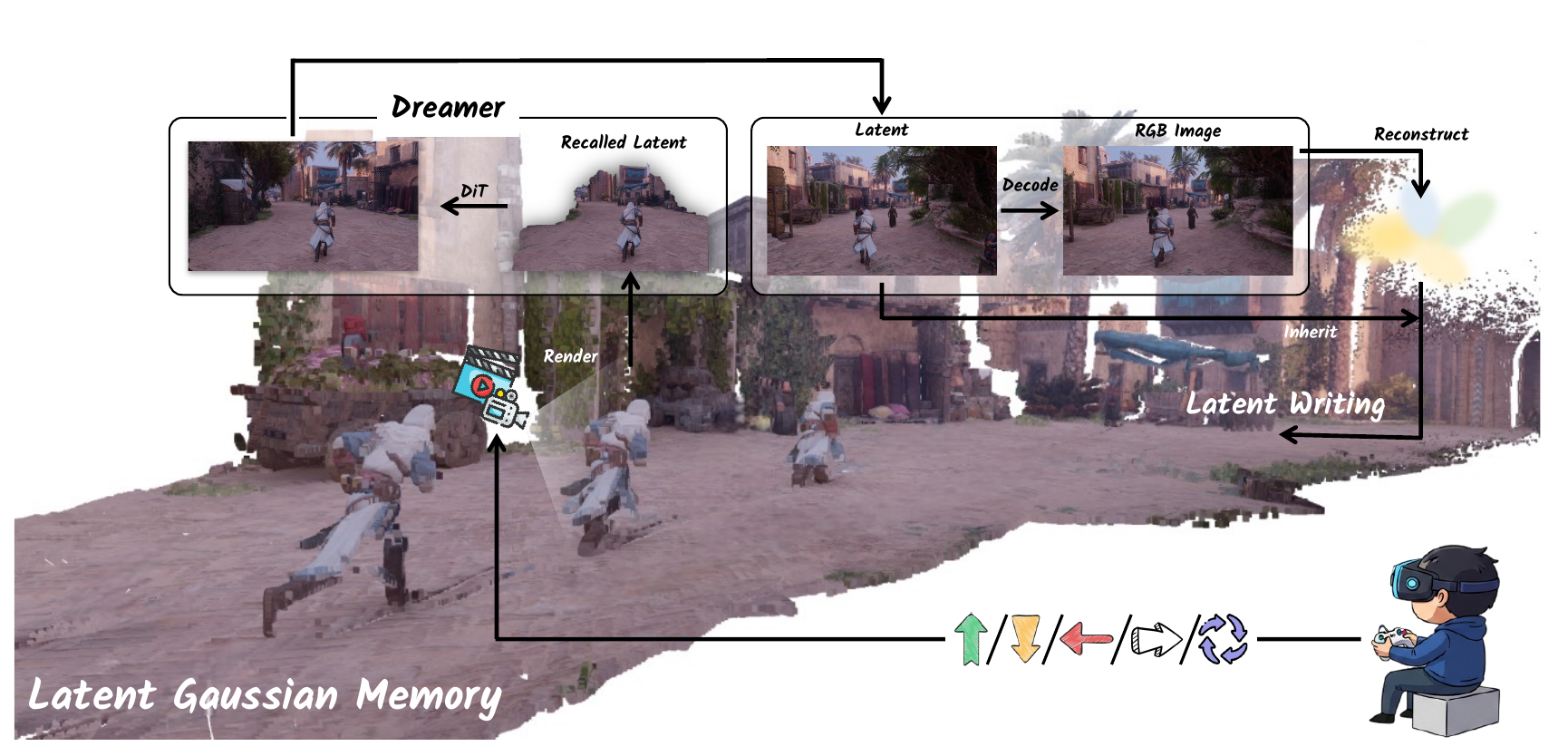}
        \captionof{figure}{\textbf{Overview of the inference pipeline.} Our system performs long-horizon frame-by-frame generation through a closed-loop autoregressive process. First, a user action triggers memory recall via Gaussian Splatting, rendering a viewpoint-aligned latent. This latent conditions the proposed Dreamer to generate the next frame, which is subsequently decoded into RGB. Finally, the generated latent is directly inherited by the reconstructed primitives to update the memory.}
        \label{fig:teaser}
    \end{minipage}
\end{center}
    \begin{abstract}
    Frame-wise action-controlled image-to-video generation is a promising paradigm for interactive world simulation, where each control signal should elicit an immediate visual response. However, maintaining visual fidelity and 3D consistency over long autoregressive rollouts remains challenging. Existing 3D-aware methods often suffer from catastrophic drift due to two impediments: information loss from \textit{Latent--RGB Cycling}, where generated latents are repeatedly decoded to RGB and re-encoded for future conditioning, and the training--inference gap induced by the \textit{error-free hypothesis}, where clean training memory fails to match prediction-corrupted inference memory. To address these challenges, we present \textbf{Robust Dreamer}, a memory-augmented framework built around how to design 3D memory and how to use it robustly. First, we introduce \textbf{Latent Gaussian Memory}, which anchors diffusion latents inherited from the generation process to Gaussian primitives and recalls them via latent-space Gaussian splatting. This provides dense, geometry-aware, view-aligned conditioning while avoiding accumulated degradation from repeated VAE conversion. Second, we propose \textbf{Deviation Learning with Dynamic Deviation Archive}, which synthesizes rollout-induced latent deviations through a one-step approximation, stores them by autoregressive stage and denoising timestamp, and injects them into historical memory during training. This exposes the generator to realistic corrupted memory states and teaches internal correction before inference. Experiments on ScanNet, DL3DV, and OmniWorldGame demonstrate state-of-the-art long-horizon performance. Project page is available at \url{https://hlinchen.github.io/projects/Robust_Dreamer/}.
\end{abstract}

\section{Introduction} \label{sec:intro}

    World simulation has gained significant attention for its potential to model complex environments and predict the outcomes of actions. By rolling out plausible future trajectories, such models enable agents to plan, reason, and learn without expensive real-world interactions. Recent advances in video diffusion models have further propelled this field, offering high-fidelity frame synthesis and strong temporal coherence over short horizons~\cite{oasis2024,wan2025,google2025veo3}. However, extending these models from short clips to consistent and interactive long-horizon rollouts remains a formidable challenge. As the model recursively predicts future frames over long but finite horizons, errors accumulated over sequential steps often lead to severe degradation in visual quality and spatial consistency~\cite{liu2021infinite, genwarp}.
    
    Existing approaches to long-term autoregressive video generation generally fall into two categories. The first relies on 2D keyframe conditioning~\cite{li2025vmem,xiao2025worldmem,yu2025context-memory}, employing attention or retrieval mechanisms to identify historical frames. While effective for stylistic coherence, these geometry-agnostic methods lack pixel-level supervision, often resulting in 3D inconsistencies and artifacts during significant viewpoint changes. The second category incorporates explicit 3D representations by lifting historical frames into a persistent 3D memory to condition generation~\cite{huang2025voyager,wu2025spmem}. By organizing historical observations in a spatially grounded representation, 3D memory provides geometry-aware constraints for future views and offers a promising path toward more consistent long-horizon generation.
    
\begin{figure}[t]
        \centering
        \includegraphics[width=0.65\textwidth]{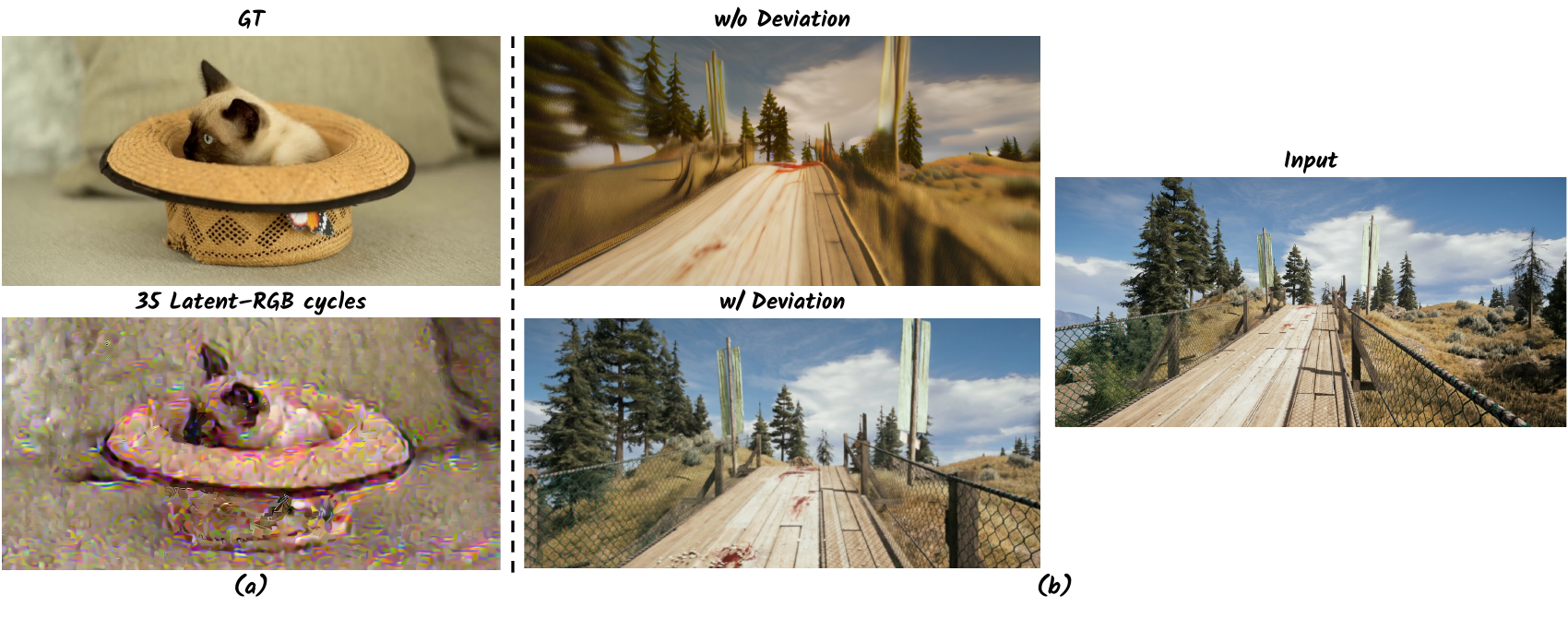}
        \captionof{figure}{
        \textbf{Motivation.} (a) \textbf{Latent--RGB Cycling:} Repeatedly decoding a latent to RGB and encoding the RGB back to latent for 35 iterations causes catastrophic signal degradation and color distortion due to accumulated quantization errors. (b) \textbf{Deviation Learning:} The baseline (top), trained on clean memory, i.e., memory constructed from clean training frames/latents, suffers from structural collapse due to the training--inference gap. In contrast, our method (bottom) explicitly models accumulated errors, effectively mitigating drift and preserving geometric details.
        }
        \vspace{-0.2cm}
        \label{fig:motivation}
    \end{figure}

    In this work, we aim to apply such 3D memory to \textit{frame-wise action-controlled image-to-video generation}, where an initial image and step-wise controls drive an interactive rollout with low-latency feedback. This task is practical for embodied agents and interactive world simulation: each control signal should trigger an immediate visual response, rather than committing one control to a long video chunk of dozens of frames. Here, 3D memory is especially important: every generated frame is written into a spatial memory and recalled to condition future actions from new viewpoints. However, we identify two key impediments in current 3D-aware frameworks. First, most methods store or render RGB observations before re-encoding them into the video model's latent space, causing \textit{Latent--RGB Cycling} and discarding fine latent details. While a single VAE decode--encode cycle may be tolerable, frame-wise generation repeats this operation at every memory update and recall, amplifying small reconstruction errors into color drift and structural degradation, as shown in \figref{fig:motivation} (a). Second, current training often assumes clean ground-truth memory, whereas inference builds memory from imperfect model predictions with accumulated artifacts. This \textit{error-free hypothesis}~\cite{huang2025self-forcing} creates a feedback loop: prediction deviations contaminate memory, corrupted memory biases future predictions, and the rollout gradually collapses, as shown in \figref{fig:motivation} (b).
    
    To address these issues, we propose \textbf{Robust Dreamer}, a memory-augmented framework for frame-wise action-controlled image-to-video generation. Our solution is organized around how to design 3D memory and how to use it robustly. For memory design, inspired by latent-level manipulation in Wan-Move~\cite{chu2025wan}, we introduce \textbf{Latent Gaussian Memory}, which treats reconstruction as an incremental memory update in the diffusion latent space. Instead of decoding generated latents into an RGB proxy for storage, we inherit latent features directly from the generation process and anchor them to Gaussian primitives whose geometry is predicted by a feed-forward reconstruction module. During recall, Gaussian splatting aggregates these inherited latents with alpha-blending weights into the target viewpoint, producing a dense, pixel-aligned latent condition for the next frame. This keeps memory write and recall inside the high-dimensional latent space, preserves fine-grained 3D context, and avoids accumulated degradation from repeated VAE conversion. For memory usage, we introduce \textbf{Deviation Learning with Dynamic Deviation Archive}. Rather than assuming an idealized memory during training, we expose the generator to realistic corrupted historical contexts. To make this feasible without expensive autoregressive unrolling, we synthesize rollout-induced latent deviations with a one-step approximation and store them in a \textbf{Dynamic Deviation Archive} hierarchically indexed by autoregressive stage and denoising timestamp. By sampling deviations from the archive and injecting them into historical memory states, the generator learns internal correction under the same type of imperfect memory it will encounter during autoregressive inference.
    
    In summary, our contributions are: 
    \begin{itemize}
        \item We propose \textbf{Latent Gaussian Memory}, a geometry-aware 3D memory that anchors inherited diffusion latents to Gaussian primitives and recalls them by splatting, enabling dense, view-aligned conditioning without repetitive latent--RGB cycling.
        \item We introduce \textbf{Deviation Learning with Dynamic Deviation Archive}, which synthesizes and stores realistic rollout-induced deviations to train the generator under corrupted memory states, bridging the training--inference gap and mitigating accumulated drift.
        \item We conduct comprehensive experiments on ScanNet, DL3DV, and OmniWorldGame, demonstrating state-of-the-art long-horizon performance, improved 3D consistency, and stronger robustness in frame-wise action-controlled video generation.
    \end{itemize}

\section{Related Work}
\label{sec:related}

\noindent\textbf{Camera-Controlled Video Generation.}
Camera-controlled video generation conditions a video model on desired viewpoint changes. Early methods directly inject camera poses, motion vectors, or ray-based embeddings into diffusion backbones~\cite{wang2024motionctrl,he2024cameractrl,bahmani2024ac3d,popov2025camctrl3d,li2025realcam}. Recent action-controlled world models can also be viewed through this lens: keyboard or user actions are often translated into camera pose trajectories that drive interactive scene exploration~\cite{genie3,li2025hunyuan-gamecraft,zhang2025matrix-game,he2025matrix,sun2025worldplay,shen2026inspatio}. While pose-level conditioning improves controllability, numerical camera signals alone provide weak spatial constraints, especially under large viewpoint changes and revisits. To improve geometric fidelity, another line conditions generation on 3D-aware signals, such as depth-warped images, point-cloud renderings, or updatable spatial representations~\cite{yu2024viewcrafter,ren2025gen3c,yu2025trajectorycrafter,li2025magicworld,zhao2025spatia}. Our method follows this geometry-aware camera-control direction, but differs in two aspects: we generate frame by frame so that interaction can be injected at every step instead of committing to a long preplanned chunk, and we use the 3D representation as a persistent latent memory rather than an RGB proxy, together with deviation-aware training to make this memory robust during autoregressive rollout.

\noindent\textbf{Memory for Video Generation.}
Autoregressive video generation requires mechanisms beyond a short sliding context. One line of work uses 2D memory, retrieving historical frames, visual tokens, or compressed context slots according to view overlap or temporal relevance~\cite{yu2025context-memory,xiao2025worldmem,oshima2025worldpack,hong2025relic,zhang2025framepack}. Such memories are lightweight and compatible with pretrained video models, but they remain view-dependent and provide limited pixel-level geometric correspondence under large camera motion. Another line builds 3D memory with point clouds, surfels, spatial maps, or geometry-aware reconstruction to support revisit consistency~\cite{huang2025voyager,wu2025geometry-forcing,li2025vmem,wu2025spmem,li2025magicworld,lu2026see4d,kong2025worldwarp,zhao2025spatia}. These methods are closer to our goal, yet most store or render RGB observations before re-encoding them into the generator, which accumulates latent--RGB cycling artifacts during autoregressive rollout. In contrast, our Latent Gaussian Memory stores inherited diffusion latents on Gaussian primitives and recalls them by splatting, preserving dense 3D-aligned context without repeated VAE conversion.

\noindent\textbf{Train--Test Gap in Autoregressive Video Generation.}
Autoregressive video diffusion enables streaming long-horizon generation~\cite{yin2025causvid,teng2025magi-1,yesiltepe2025infinity-rope}, but training on ground-truth histories and testing on self-generated histories creates exposure bias and accumulated drift. The forcing family studies this mismatch by training or sampling under rollout-like conditions: Diffusion Forcing and history-guided diffusion model partially observed sequences~\cite{chen2025diffusion,song2025historyguidedvideodiffusion}, Self-Forcing and Self-Forcing++ fine-tune on generated contexts~\cite{huang2025self-forcing,cui2025self-forcing2}, while Rolling Forcing and self-resampling inject noisy or resampled histories to improve robustness~\cite{liu2025rolling,guo2025resampling}. Causal Forcing further identifies an architectural gap in distilling bidirectional diffusion models into causal AR students and uses an AR teacher for ODE initialization to improve real-time interactive generation~\cite{zhu2026causal}. Stable Video Infinity studies error-bank recycling for very long rollouts~\cite{li2025svi}, and recent analyses explicitly characterize error accumulation in AR video diffusion~\cite{wang2025error}. These approaches mostly address generic 2D histories or output frames. Our Deviation Learning targets the geometry-aware memory setting: it synthesizes realistic latent deviations and stores them in a Dynamic Deviation Archive, so the model learns to generate from corrupted memory states similar to those encountered at inference.

    \section{Robust Dreamer: Memory-Augmented Generation}
\label{sec:method}

In this work, we address action-controlled autoregressive video generation. Given an initial frame \(\bm{I}_0\) and a step-wise control signal \(c_i\), the model predicts the next frame \(\bm{I}_{i+1}\) from the current rollout state. Our framework is built around two complementary memories. The \textbf{Latent Gaussian Memory} \(\mathcal{M}\) provides geometrically grounded long-range context by storing generated content on Gaussian primitives, avoiding repeated latent--RGB conversion. The \textbf{Dynamic Deviation Archive} \(\mathcal{A}\) models the errors accumulated by autoregressive rollout and exposes the generator to such corrupted memory during training. Sec.~\ref{sec:framework} gives the full training and inference pipeline, Sec.~\ref{sec:latent_memory} details the latent Gaussian memory, and Sec.~\ref{sec:deviation_archive} presents Deviation Learning and the archive mechanism.

\begin{figure*}
    \centering
    \includegraphics[width=0.95\textwidth]{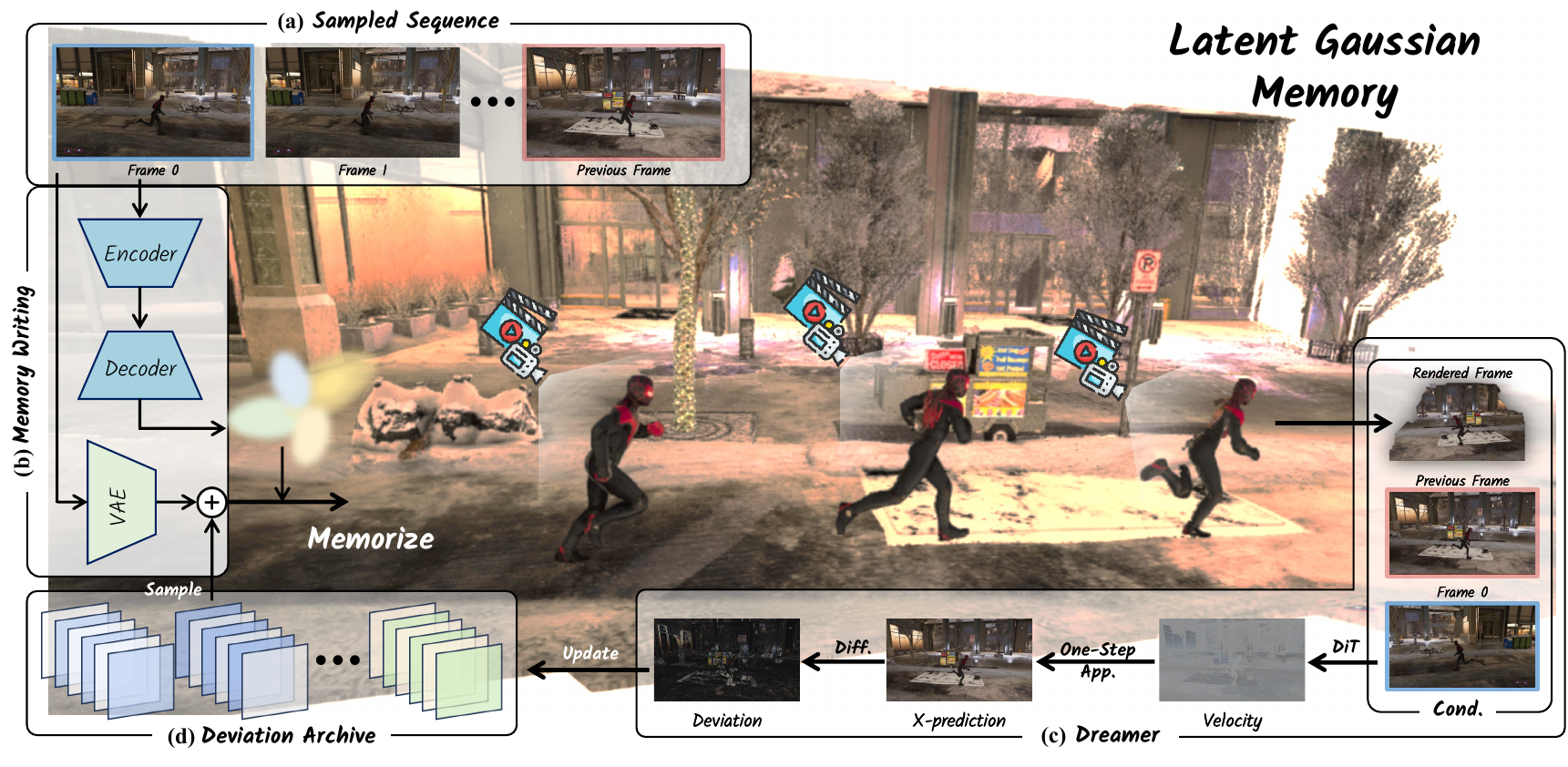}
    \captionof{figure}{
    \textbf{Overview of the training pipeline.} (a) Variable-length subsequences provide historical context. (b) \textbf{Latent Gaussian Memory} is built from deviation-corrupted histories. (c) The Dreamer predicts velocity conditioned on clean anchor (frame 0), predecessor (previous frame), and recalled latent memory (rendered frame). (d) One-step deviations update the \textbf{Dynamic Deviation Archive}.
    }
    \label{fig:training}
\end{figure*}

\subsection{Memory-Conditioned Rollout Framework}
\label{sec:framework}

We describe the overall generation and training protocol. We operate in the latent space of a pretrained video diffusion model and denote the latent of \(\bm{I}_i\) as \(\bm{X}_i\). The idea is to use a persistent latent memory to provide geometrically aligned long-range context during rollout, while training the generator with archived deviations that approximate inference-time errors.

\noindent\textbf{Memory-Conditioned Autoregressive Inference.}
At inference time, Robust Dreamer maintains the Latent Gaussian Memory as an online state. The memory is initialized from the clean input pair \((\bm{I}_0,\bm{X}_0)\). At generation step \(i\), given the control \(c_i\), we render the memory from the target viewpoint and obtain a view-aligned latent condition \(\hat{\bm{X}}_{i}^{\mathcal{M}}\) for predicting \(\bm{X}_{i+1}\). The Dreamer is conditioned on four complementary signals: the clean anchor \(\bm{X}_0\), the latest rollout latent \(\bm{X}_i\), the recalled memory feature \(\hat{\bm{X}}_{i}^{\mathcal{M}}\), and the control \(c_i\):
\begin{equation}
    \bm{y}_i =
    \texttt{concat}(\bm{X}_0, \bm{X}_i, \hat{\bm{X}}_{i}^{\mathcal{M}}, c_i).
    \label{eq:overview_infer_condition}
\end{equation}
The DiT denoiser predicts the next latent \(\bm{X}_{i+1}\) from noise under this condition, and the latent is decoded into the RGB frame \(\bm{I}_{i+1}\). The generated pair \((\bm{I}_{i+1},\bm{X}_{i+1})\) is then written into the Latent Gaussian Memory, providing the geometric state for the subsequent control \(c_{i+1}\).

\noindent\textbf{Deviation-Aware Training.}
Training with clean ground-truth histories does not match the inference regime, where the memory is accumulated from imperfect predictions. To expose the generator to this rollout-induced distribution shift, we train the Dreamer with deviation-corrupted memory conditions. Given a sampled subsequence \(\{\bm{X}_0,\dots,\bm{X}_{n+1}\}\), the first frame serves as the clean anchor and the final frame \(\bm{X}_{n+1}\) is used as the clean target. We keep the anchor \(\bm{X}_0\) uncorrupted, matching the practical setting where the initial frame is provided by the user, while subsequent frames are generated by the model and thus susceptible to accumulated drift. For each historical frame \(1 \le i \le n\), we sample a deviation \(\bm{D}_i\) from the Dynamic Archive \(\mathcal{A}\) and inject it probabilistically:
\begin{equation}
    \tilde{\bm{X}}_i = \bm{X}_i + \mathbb{I}_i \bm{D}_i,
    \quad \mathbb{I}_i \sim \mathrm{Bernoulli}(p).
    \label{eq:overview_deviation_inject}
\end{equation}
Unlike inference, where memory is updated online after each generated frame, training constructs a temporary corrupted memory \(\tilde{\mathcal{M}}\) from the sampled history \((\bm{X}_0,\tilde{\bm{X}}_1,\dots,\tilde{\bm{X}}_{n})\). We render this memory for the target viewpoint to obtain \(\hat{\bm{X}}_{n}^{\tilde{\mathcal{M}}}\) and form the training condition
\begin{equation}
    \bm{y}_n =
    \texttt{concat}(\bm{X}_0,\tilde{\bm{X}}_{n},\hat{\bm{X}}_{n}^{\tilde{\mathcal{M}}},c_{n}).
    \label{eq:overview_train_condition}
\end{equation}
The supervision remains the clean future latent \(\bm{X}_{n+1}\). With a noisy latent \(\bm{X}_{n+1}^t=t\bm{X}_{n+1}+(1-t)\bm{\epsilon}\), where \(\bm{\epsilon} \sim\mathcal{N}(0,\mathbf{I})\), we optimize the flow-matching objective to regress the velocity field
\begin{equation}
    \mathcal{L} =
    \mathbb{E}_{t}\left[
    \left\|
    \mathrm{DiT}(\bm{X}_{n+1}^t,\bm{y}_n,t;\theta)
    - (\bm{X}_{n+1}- \bm{\epsilon})
    \right\|_2^2
    \right].
    \label{eq:overview_loss}
\end{equation}
We train the generator to denoise the future latent from deviation-corrupted history and memory recall. The archive \(\mathcal{A}\) is updated online with one-step synthesized deviations, as detailed in Sec.~\ref{sec:deviation_archive}.

\subsection{Geometry-Grounded Latent Gaussian Memory}
\label{sec:latent_memory}

Action-controlled autoregressive video generation requires a memory that is both geometrically persistent and compatible with the latent space of the video generator. A common design in existing 3D-aware memories is to lift historical RGB frames into point-cloud-like 3D representations and later project them for future conditioning~\cite{huang2025voyager,li2025vmem,wu2025spmem}. While this provides spatial grounding, the stored content repeatedly crosses the latent--RGB interface of the VAE, causing quantization and detail loss. Point-based projection is also sparse under novel views, leaving holes and disocclusion artifacts. Inspired by 3D Gaussian Splatting~\cite{kerbl20233dgs} and latent-space motion transfer~\cite{chu2025wan}, we instead store generated diffusion latents on continuous Gaussian primitives and recall them directly through alpha-composited splatting in latent space.

\noindent\textbf{Latent Gaussian Memory.} Formally, after processing frames up to frame \(i\), we define the memory as a set of Gaussian primitives
\begin{equation}
    \mathcal{M}_i = \{ \mathbf{g}_s \}_{s=1}^{S_i}, \quad
    \mathbf{g}_s = (\bm{\mu}_s, o_s, \bm{q}_s, \bm{s}_s, \bm{x}_s),
    \label{eq:latent_memory_bank}
\end{equation}
where \(\bm{\mu}_s \in \mathbb{R}^3\) is the center of the primitive in a global coordinate system, and \(o_s\), \(\bm{q}_s\), and \(\bm{s}_s\) denote its opacity, rotation, and anisotropic scale, respectively. The last attribute \(\bm{x}_s \in \mathbb{R}^C\) is the latent feature stored by the primitive. This parameterization decouples geometry from content: the Gaussian attributes determine where and how a memory element is projected, while \(\bm{x}_s\) carries the semantic and appearance information inherited from the generated latent. For dynamic scenes, we further augment each primitive with time-dependent Gaussian attributes following 4D Gaussian representations~\cite{yang2023real}, enabling a unified interface for static 3D recall and temporal 4D recall.

\noindent\textbf{Pixel-Aligned Latent Writing.}
At generation step \(i\), when the frame \(\bm{I}_i\) becomes available, we write its latent \(\bm{X}_i\) into the memory. We use a feed-forward reconstruction backbone initialized from CUT3R-style geometry prediction~\cite{wang2025cut3r,wang2024dust3r,mast3r_arxiv24}, to align the current observation with the accumulated global scene. Concretely, \(\bm{I}_i\) is encoded into visual tokens using a ViT encoder~\cite{dosovitskiy2020image} to obtain context-aware image tokens \(\bm{F}'_i\) and a pose token \(\bm{z}'_i\). 
On top of these context-aware tokens, we train DPT prediction heads to output the geometry and Gaussian attributes required by the memory bank:
\begin{equation}
    \hat{\bm{P}}_i = \texttt{DPT}_{\mathrm{global}}(\bm{F}'_i, \bm{z}'_i), \quad
    \{(\hat{o}_{ij}, \hat{\bm{q}}_{ij}, \hat{\bm{s}}_{ij})\}_{j=1}^{HW}
    = \texttt{DPT}_{\mathrm{3DGS}}(\bm{F}'_i, \bm{z}'_i).
    \label{eq:memory_write_heads}
\end{equation}
The global head follows the CUT3R-style world pointmap prediction and produces per-pixel 3D points \(\hat{\bm{P}}_i\) in the global coordinate system, which serve as Gaussian centers \(\bm{\mu}_{ij}\). The 3DGS head predicts the Gaussian parameters for each pixel-anchored primitive. To assign latent content, we first resize the generated latent map \(\bm{X}_i\) to the image resolution and then read \(\bm{x}_{ij}\) from the same pixel location as the corresponding primitive. Thus, memory writing lifts generated latents into globally aligned Gaussian primitives without predicting or re-encoding their semantic content.

For dynamic scenes, we use an additional 4DGS head to predict time-dependent rotation and scale for moving content. In practice, we also control the primitive count by downsampling the prediction resolution and merging nearby primitives through voxel-based pruning inspired by Scaffold-GS~\cite{scaffoldgs}. We provide these details in the supplementary material (\secref{sec:mem_compress} and \secref{sec:mem_ret}).

\noindent\textbf{Splatting-Based Latent Recall.}
To condition generated frame \(i\), we query the memory from the target viewpoint specified by the control \(c_i\). Rather than rendering an RGB image and re-encoding it, we directly splat the latent features stored in \(\mathcal{M}_i\) into the target view. For a pixel \(k\) in the recalled latent map, let \(\mathcal{N}(k)\) be the set of memory primitives whose projected Gaussians overlap that pixel, sorted by depth. The recalled feature is computed with the standard alpha-compositing form
\begin{equation}
    \hat{\bm{X}}^{\mathcal{M}}_{i}(k)
    = \sum_{j \in \mathcal{N}(k)} T_j w_j \bm{x}_j,
    \quad
    T_j = \prod_{m < j} (1 - w_m),
    \label{eq:latent_memory_recall}
\end{equation}
where \(w_j\) is the splatting weight of the \(j\)-th primitive at pixel \(k\), and \(T_j\) is the accumulated transmittance before that primitive. For static content, \(w_j\) is determined by the projected 2D Gaussian density and opacity; for dynamic content, it is computed by slicing the corresponding 4D Gaussian at the target timestamp. We provide these retrieval details in \secref{sec:mem_ret} in the supplementary material. The resulting dense latent map \(\hat{\bm{X}}^{\mathcal{M}}_{i}\) is spatially aligned with the requested viewpoint and is used as a structural condition for predicting \(\bm{X}_{i+1}\), closing the write--recall loop of autoregressive generation.

\subsection{Deviation Learning with Dynamic Deviation Archive}
\label{sec:deviation_archive}

Autoregressive generation is vulnerable to accumulated deviation. During inference, each generated frame is written into the Latent Gaussian Memory and later recalled as part of the condition for future steps. Consequently, even small prediction errors are not isolated: they contaminate the memory, bias subsequent denoising, and are repeatedly reintroduced through the write--recall loop. This creates a training--inference mismatch, because standard training constructs memory from clean ground-truth histories, whereas inference constructs memory from the model's own imperfect outputs. A faithful solution would simulate full autoregressive rollouts during training, but doing so with iterative diffusion sampling is prohibitively expensive. We therefore introduce \textbf{Deviation Learning}: instead of replaying complete rollouts, we maintain a \textbf{Dynamic Deviation Archive} that stores realistic latent deviations synthesized online and reuses them to corrupt training histories.

\noindent\textbf{Archive Structure.}
The archive is designed to approximate the non-stationary deviation distribution induced by the current generator. Deviations vary with the autoregressive stage, since errors at later frames contain longer accumulated history. They also depend on the denoising timestamp at which they are synthesized, because velocity prediction errors at different noise levels exhibit different deviation patterns. We therefore organize the archive with two levels: the frame index \(i\) of the autoregressive rollout and the denoising timestamp \(t\):
\begin{equation}
    \mathcal{A} = \{\mathcal{A}_i\}_{i=1}^{N}, \quad
    \mathcal{A}_i = \{\mathcal{A}_{i}^{t}\}_{t \in \mathcal{T}}, \quad
    \mathcal{A}_{i}^{t} = \{\bm{D}_{i,t,z}\}_{z=1}^{Z}.
    \label{eq:deviation_archive}
\end{equation}
Here \(\mathcal{A}_i\) is the frame-specific archive for the \(i\)-th generated frame, \(\mathcal{A}_i^t\) is the cell associated with denoising timestamp \(t\), and \(z\) indexes the \(Z\) deviation slots stored in that cell. Each entry \(\bm{D}_{i,t,z} \in \mathbb{R}^{C \times H \times W}\) stores a latent-space deviation between an approximate model prediction and the corresponding clean latent at frame \(i\). This structure lets the archive preserve diverse error modes across both rollout depth and denoising state: early-stage entries capture local appearance or pose perturbations, whereas late-stage entries include accumulated color drift, geometric distortion, and semantic inconsistency.

\noindent\textbf{One-Step Deviation Synthesis.}
A direct way to obtain deviations is to run the complete diffusion sampler and compare the generated latent with the ground truth. This is too expensive to perform for every training sample. Instead, we synthesize deviations with a one-step approximation under the current model. Given a noised latent \(\bm{X}_i^t=t\bm{X}_i+(1-t)\bm{\epsilon}\) and the conditioning \(\bm{y}_{i-1}\) used to predict \(\bm{X}_i\), the DiT predicts a velocity
\begin{equation}
    \hat{\bm{V}}_i^t =
    \mathrm{DiT}(\bm{X}_i^t,\bm{y}_{i-1},t;\theta).
    \label{eq:deviation_velocity}
\end{equation}
Following the flow-matching trajectory, we approximate the terminal latent with a single Euler step toward the data endpoint and define the synthesized deviation as
\begin{equation}
    \bar{\bm{X}}_i = \bm{X}_i^t + (1-t)\hat{\bm{V}}_i^t,
    \quad
    \bm{D}_i = \bar{\bm{X}}_i - \bm{X}_i.
    \label{eq:one_step_deviation}
\end{equation}
This computation is performed without gradient updates and serves only to estimate the deviation patterns induced by the current generator. 

\noindent\textbf{Diversity-Preserving Refresh.}
During training, each synthesized deviation is inserted into the archive cell matching its autoregressive frame and denoising timestamp. Thus, a deviation \(\bm{D}_i\) synthesized for frame \(i\) at timestamp \(t\) enters \(\mathcal{A}_{i}^{t}\). Each cell has capacity \(Z\), bounding memory usage and preventing dominance by frequent stages or timestamps.

If the target cell is not full, the new deviation is appended. Once the cell reaches capacity, we update it with a diversity-preserving replacement rule. Rather than discarding a random entry, we replace the stored deviation most similar to the new one:
\begin{equation}
    z^\star = \arg\min_{z}\|\bm{D}_i-\bm{D}_{i,t,z}\|_2,
    \quad
    \bm{D}_{i,t,z^\star} \leftarrow \bm{D}_i.
    \label{eq:archive_update}
\end{equation}
This rule removes redundant deviations while retaining rare modes such as structural warping, long-range appearance drift, or semantic inconsistency. Because the generator changes throughout training, the archive is continuously refreshed: newly synthesized deviations gradually replace stale deviations from earlier model states, allowing \(\mathcal{A}\) to track the evolving error distribution.

When constructing corrupted latent Gaussian memory during training, we sample deviations according to each historical latent's stage. For a history frame at index \(i\), we select \(\mathcal{A}_i\), sample a timestamp \(t \in \mathcal{T}\) uniformly, and draw an entry from \(\mathcal{A}_i^t\). The sampled deviation is injected into the clean latent before memory writing, so both the predecessor condition and recalled memory inherit realistic inference-like artifacts. The archive is training-only: it calibrates the Dreamer to denoise under corrupted memory conditions, while inference maintains only the Latent Gaussian Memory and accumulates deviations naturally through generated frames autoregressively.

\section{Experiments}
    \label{sec:exp}

    We evaluate our method experimentally.
    We cover settings in \secref{sec:settings}, results in \secref{sec:results}, and ablations in \secref{sec:ablation}.
    More visualizations and analyses are in Appendix \secref{sec:more_qualitative} and \secref{sec:runtime_memory_robustness}, respectively.

    \subsection{Experimental Settings}
        \label{sec:settings}
        
        \textbf{Datasets.} We evaluate our framework on three distinct datasets covering diverse scenarios, ranging from real-world indoor scenes to game-style simulation environments.
        First, we use ScanNet~\cite{scannet}, a large-scale indoor dataset, to assess the model's ability to handle complex interior geometries and textures.
        Second, we employ DL3DV~\cite{ling2024dl3dv}, which features high-quality 3D scenes with complex camera trajectories, to evaluate robustness in 3D-consistent view synthesis.
        Finally, to evaluate our method on long-horizon interactive trajectories, we utilize OmniWorldGame~\cite{zhou2025omniworld}, a dataset designed for extended world-simulation rollouts.
        
        \textbf{Evaluation metrics.} Following standard practices in video generation and novel view synthesis, we assess the performance quantitatively using four metrics covering pixel-level fidelity, perceptual quality, and distribution realism:
        (1) PSNR (Peak Signal-to-Noise Ratio) and (2) SSIM (Structural Similarity Index Measure) are used to evaluate pixel-wise accuracy and structural preservation compared to ground truth frames.
        (3) LPIPS (Learned Perceptual Image Patch Similarity) is employed to measure perceptual degradation, as it aligns better with human visual perception.
        (4) FID (Fréchet Inception Distance) is calculated to assess the overall visual realism and the distributional distance between the generated sequences and the real data.
    
    \subsection{Results}
    \label{sec:results}
        \newcolumntype{Y}{>{\centering\arraybackslash}X}

\begin{table*}[t]
    \centering
    \footnotesize
    \caption{Short-/long-term (80/300 frames) generation results on the ScanNet.}
    \vspace{-2mm}
    \label{tab:scannet}
    \renewcommand{\arraystretch}{1.2}
    \setlength{\tabcolsep}{1pt} 
    \begin{tabularx}{\textwidth}{@{} l *{7}{Y} @{\hspace{4pt}} *{7}{Y} @{}}
    \toprule
    \textbf{Metric} & \multicolumn{7}{c}{\textbf{Short-term (80 frames)}} & \multicolumn{7}{c}{\textbf{Long-term (300 frames)}} \\
    \cmidrule(r{20pt}){2-8} \cmidrule(l){9-15}
    & MCtl & CCtl & WMem & WWarp & VMem & SForc & \textbf{Ours} & MCtl & CCtl & WMem & WWarp & VMem & SForc & \textbf{Ours} \\
    \midrule
    PSNR $\uparrow$ & 12.14 & 12.35 & 14.12 & 15.76 & 13.21 & 13.95 & \textbf{16.89} & 9.14 & 9.35 & 11.12 & 11.76 & 9.21 & 9.38 & \textbf{13.43} \\
    SSIM $\uparrow$ & 0.185 & 0.198 & 0.258 & 0.312 & 0.224 & 0.246 & \textbf{0.651} & 0.105 & 0.118 & 0.158 & 0.212 & 0.124 & 0.136 & \textbf{0.422} \\
    LPIPS $\downarrow$ & 0.485 & 0.462 & 0.388 & 0.345 & 0.415 & 0.397 & \textbf{0.351} & 0.625 & 0.582 & 0.508 & 0.445 & 0.515 & 0.502 & \textbf{0.381} \\
    FID $\downarrow$ & 85.34 & 81.12 & 42.50 & 38.65 & 56.40 & 44.20 & \textbf{16.82} & 125.34 & 118.12 & 72.50 & 68.65 & 96.40 & 92.40 & \textbf{38.42} \\
    \bottomrule
    \end{tabularx}
    \vspace{-0.3cm}
\end{table*}

        We present the main experimental results of our method, evaluating its performance in both short- and long-term video generation. We compare against a diverse set of state-of-the-art baselines to assess visual quality, temporal coherence, and robustness under long-horizon rollout. Specifically, we include MotionCtrl (MCtl)~\cite{wang2024motionctrl} and CameraCtrl (CCtl)~\cite{he2024cameractrl} as representative non-autoregressive control-based methods, and Self Forcing (SForc)~\cite{huang2025self-forcing} as a recent frame-by-frame autoregressive approach designed to alleviate error accumulation. We further compare with memory-based autoregressive world modeling methods, WorldMem (WMem)~\cite{xiao2025worldmem}, WorldWarp (WWarp)~\cite{kong2025worldwarp}, and VMem~\cite{li2025vmem}, which enhance long-term consistency via explicit or implicit memory mechanisms.

\begin{figure*}[t]
            \centering
            \includegraphics[width=\textwidth]{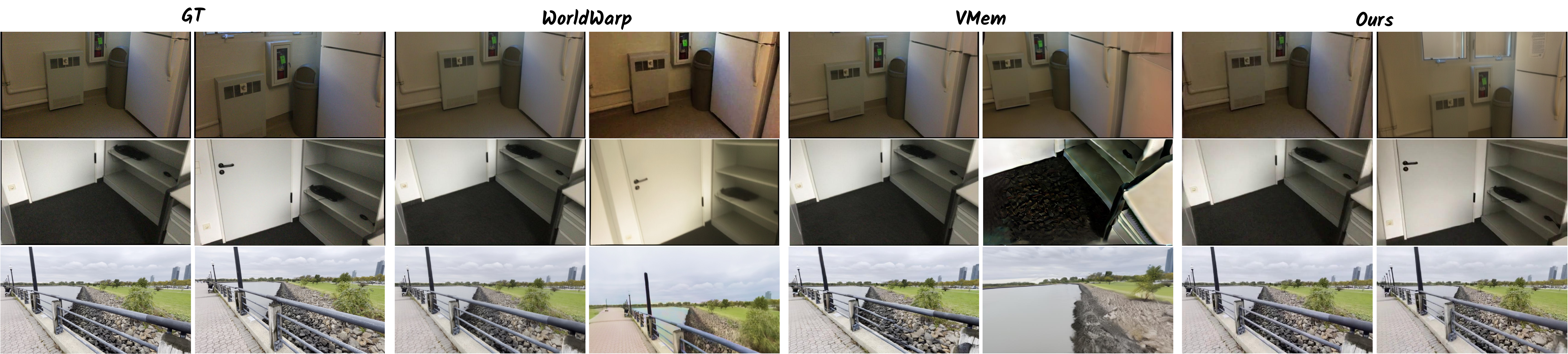}
            \captionof{figure}{
            \textbf{Qualitative results on ScanNet and DL3DV.} For each method, we visualize the first generated frame (left) and a later frame in the rollout (right). As generation progresses, baseline methods suffer from accumulated color drift and structural degradation, whereas our approach maintains consistent geometry and appearance without noticeable misalignment.
            }
            \vspace{-0.35cm}
            \label{fig:vis}
        \end{figure*}

        \textbf{Comparisons on the ScanNet Dataset.}
        ScanNet provides long and continuous indoor video sequences with complex camera motion and rich geometric structure, making it particularly suitable for evaluating long-horizon video generation. Quantitative comparisons are reported in \tabref{tab:scannet} for both short-term (80 frames) and long-term (300 frames) settings. As shown in the table, our method consistently achieves the best overall performance across all evaluation metrics. In the short-term regime, our approach maintains competitive visual fidelity, while in the long-term setting it exhibits clear advantages, achieving higher PSNR and SSIM together with lower LPIPS and FID. This indicates stronger robustness to error accumulation during extended autoregressive rollout. Qualitative results in \figref{fig:vis} further demonstrate that our method preserves coherent geometry and appearance over long sequences.

        \begin{table*}[t]
    \centering
    \caption{Quantitative comparison on the DL3DV and OmniWorldGame datasets.}
    \label{tab:dl3dv}
    \resizebox{\linewidth}{!}{%
    \begin{tabular}{lccccccc|ccccccc}
    \toprule
    & \multicolumn{7}{c|}{\textbf{DL3DV}} & \multicolumn{7}{c}{\textbf{OmniWorldGame}} \\
    \cmidrule(lr){2-8} \cmidrule(lr){9-15}
    \textbf{Metric} & \textbf{MCtl} & \textbf{CCtl} & \textbf{SForc} & \textbf{WMem} & \textbf{WWarp} & \textbf{VMem} & \textbf{Ours} & \textbf{MCtl} & \textbf{CCtl} & \textbf{SForc} & \textbf{WMem} & \textbf{WWarp} & \textbf{VMem} & \textbf{Ours} \\
    \midrule
    PSNR $\uparrow$    & 10.82 & 11.15 & 12.60 & 12.85 & 13.64 & 12.15 & \textbf{15.42} & 13.45 & 13.82 & 15.12 & 15.45 & 16.12 & 14.25 & \textbf{17.43} \\
    SSIM $\uparrow$    & 0.145 & 0.162 & 0.215 & 0.224 & 0.285 & 0.194 & \textbf{0.522} & 0.245 & 0.258 & 0.352 & 0.315 & 0.382 & 0.284 & \textbf{0.541} \\
    LPIPS $\downarrow$ & 0.524 & 0.495 & 0.418 & 0.412 & 0.385 & 0.435 & \textbf{0.291} & 0.385 & 0.364 & 0.295 & 0.345 & 0.312 & 0.354 & \textbf{0.242} \\
    FID $\downarrow$   & 92.45 & 88.12 & 53.80 & 52.34 & 48.67 & 64.20 & \textbf{22.15} & 65.20 & 62.15 & 42.12 & 45.30 & 36.85 & 52.40 & \textbf{21.92} \\
    \bottomrule
    \end{tabular}}
    \vspace{-0.1cm}
\end{table*}

        \textbf{Comparisons on the DL3DV Dataset.}
        DL3DV is a more challenging dataset that contains both indoor and outdoor scenes, with larger viewpoint changes and more diverse appearance variations. These factors introduce additional difficulty for autoregressive rollout. Quantitative results on DL3DV in \tabref{tab:dl3dv} show that our method consistently outperforms all baselines across evaluation metrics. While baseline methods exhibit noticeable degradation when transitioning between indoor and outdoor environments, our approach maintains higher visual fidelity and temporal consistency. This demonstrates that the proposed method generalizes well across heterogeneous scene types and remains robust under more challenging real-world conditions.

        \textbf{Comparisons on the OmniWorldGame Dataset.}
        OmniWorldGame features complex dynamic scenes with non-trivial object motion and interaction. Compared to static or quasi-static real-world datasets, OmniWorldGame poses additional challenges for maintaining temporal coherence in the presence of dynamic content. As shown in \tabref{tab:dl3dv}, our method achieves the best overall performance. Baseline methods tend to suffer from motion drift or temporal inconsistency as dynamics accumulate, whereas our approach preserves both scene structure and dynamic behavior over extended horizons. These results highlight the effectiveness of our method in handling dynamic environments.
    
    \subsection{Ablation Study}
        \label{sec:ablation}
        \begin{figure}[t]
    \centering
    \begin{minipage}[t]{0.55\linewidth}
        \centering
        \vspace{0pt}
        \captionof{table}{Ablation study on different components.}
        \label{tab:ablation}
        \small
        \setlength{\tabcolsep}{2pt}
        \resizebox{\linewidth}{!}{%
        \begin{tabular}{@{}lcccc@{}}
        \toprule
        Method & \textbf{PSNR} $\uparrow$ & \textbf{SSIM} $\uparrow$ & \textbf{LPIPS} $\downarrow$ & \textbf{FID} $\downarrow$ \\
        \midrule
        A. w/o. Latent       & 15.92  & 0.524  & 0.402  & 24.85 \\
        B. w/o. Deviation    & 14.10  & 0.268  & 0.392  & 48.20 \\
        C. w/ Gaussian Noise & 14.25  & 0.275  & 0.385  & 47.50 \\
        D. Ours Full         & \textbf{16.89} & \textbf{0.651} & \textbf{0.351} & \textbf{16.82} \\
        \bottomrule
        \end{tabular}
        }
    \end{minipage}
    \hfill
    \begin{minipage}[t]{0.40\linewidth}
        \centering
        \vspace{0pt}
        \includegraphics[width=\linewidth]{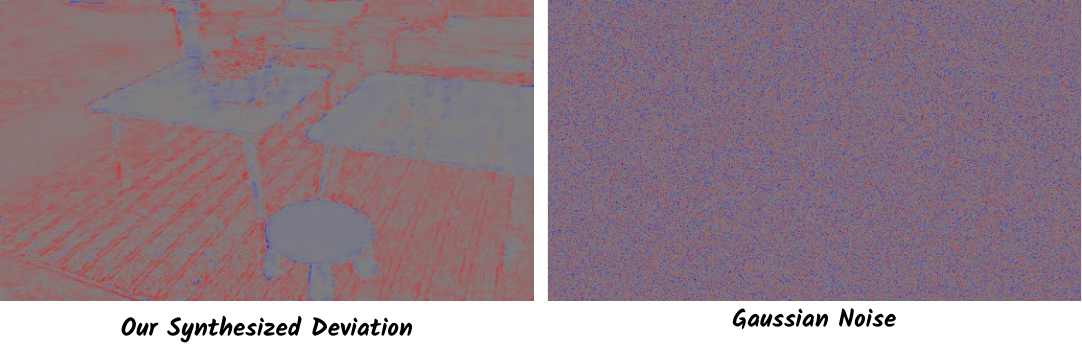}
        \vspace{-0.35cm}
        \captionof{figure}{
        \textbf{Deviation patterns.}
        Comparison between our synthesized deviation (left) and Gaussian noise (right).
        }
        \label{fig:vis_dev}
    \end{minipage}
    \vspace{-0.4cm}
\end{figure}

        We conduct comprehensive ablation studies on the ScanNet dataset~\cite{scannet} to validate the effectiveness of our individual modules. The quantitative results are in \tabref{tab:ablation}.
        
        \textbf{Effectiveness of Latent Gaussian Memory (Row A).} We first analyze the impact of performing memory writing and recall directly in the latent space with our Latent Gaussian Memory. We compare our full model (Row D) against a variant where the memory operates in the RGB pixel space (Row A, ``w/o. Latent''). In this setting, recalled content is decoded to RGB and then re-encoded via VAE for the next step. As shown in Row A, this \textit{Latent--RGB Cycling} leads to a significant performance drop across all metrics due to repeated quantization errors. This confirms that anchoring and inheriting features directly in the latent space is crucial for preserving high-frequency details.
        
        \textbf{Impact of Deviation Learning (Row B).} Next, we investigate the necessity of our Deviation Learning strategy in bridging the training--inference gap. The baseline trained solely on clean, ground-truth memory (Row B, ``w/o. Deviation'') exhibits poor performance. Without exposure to imperfect historical contexts during training, Model B lacks the capability to correct accumulated errors at inference time, leading to catastrophic drift. This result validates that explicitly training on error-corrupted memory states is essential for robust long-term generation.
        
        \textbf{Validity of One-Step Deviation Synthesis (Row C).} To verify whether our synthesized deviation captures realistic error patterns, we replace our one-step deviation synthesis with simple additive Gaussian noise (Row C, ``w/ Gaussian Noise''). The results in Row C show that training with simple noise yields suboptimal performance compared to our full method (Row D). This suggests that the distribution shift caused by autoregressive generation is structural and data-dependent, rather than random. Our proposed deviation synthesis successfully models these specific artifacts, enabling the model to learn effective internal correction. A visual comparison is provided in \figref{fig:motivation} (b).
    \textbf{Visualization of Deviation Patterns.} To further illustrate the difference between our synthesized deviation and random noise, we visualize the deviations decoded into RGB space in \figref{fig:vis_dev}. Unlike standard Gaussian noise which appears as uniform static, our synthesized deviation exhibits structure-aware patterns, such as edge blurring, that closely mirror the actual artifacts observed during inference.

    \section{Conclusion}
    \label{sec:conclusion}

    In this paper, we presented \textbf{Robust Dreamer}, a memory-augmented framework for long-horizon world simulation that addresses the critical challenges of Latent--RGB Cycling and the training--inference gap. We introduced \textbf{Latent Gaussian Memory} to eliminate repeated latent--RGB conversion by anchoring inherited diffusion latents to Gaussian primitives and recalling them directly through latent-space splatting. We further proposed \textbf{Deviation Learning with Dynamic Deviation Archive} to robustify the Dreamer against accumulated drift by synthesizing rollout-induced latent deviations and injecting them into historical memory during training. Extensive experiments on ScanNet, DL3DV, and OmniWorldGame demonstrate that our approach significantly outperforms SOTA methods, maintaining high visual fidelity and strong 3D consistency over long horizons.

    \clearpage
    {
    \small

    \bibliographystyle{plain}

    \bibliography{main}
    }

    \clearpage
    
\appendix

\section{Memory Compression Strategy.}
    \label{sec:mem_compress}
    Standard feed-forward 3DGS methods typically assign one Gaussian per pixel. While feasible for sparse-view settings, this approach becomes intractable for our frame-by-frame autoregressive generation. For instance, a standard 81-frame sequence at $512 \times 288$ resolution would yield over 11 million Gaussians, creating an excessive burden on both memory storage and retrieval efficiency. To maintain a compact and efficient memory representation, we employ a dual-level pruning strategy.
    
    First, at the 2D level, we recognize that per-pixel assignment leads to spatial redundancy; thus, we apply a $2\times$ downsampling to the prediction resolution, reducing the initial primitive count by $75\%$. Second, at the 3D level, since autoregressive generation often reconstructs overlapping geometric surfaces across adjacent frames, simply accumulating primitives results in significant redundancy. To resolve this, we introduce a Voxel-based Pruning mechanism inspired by \cite{scaffoldgs} to fuse spatially adjacent primitives into a unified representation.
    
    Specifically, we discretize the spatial domain into a grid with voxel size \(\epsilon\). For each of the \(N\) generated Gaussians with centers \(\{\bm{\mu}_j\}_{j=1}^N\), we determine its corresponding voxel index \(\boldsymbol{v}_j\): $\boldsymbol{v}_j \;=\; \left\lfloor \frac{\bm{\mu}_j}{\epsilon} \right\rceil$.
    Let \(\mathcal{G}_s = \{j \mid \boldsymbol{v}_j = s\}\) denote the set of Gaussians falling into voxel \(s\). To ensure the pruning process remains end-to-end differentiable, we predict a confidence score \(c_j\) for each Gaussian and compute intra-voxel contribution weights \(m_j\) via a local softmax: \small{$m_j =\frac{\exp(c_j)}{\sum_{k \in \mathcal{G}_s}\exp(c_k)}, \quad \text{for } j \in \mathcal{G}_s$}.
    Finally, all Gaussian attributes \(a_j\), including opacity, scale, rotation, and crucially, the inherited latent feature $\bm{x}_j$, are aggregated into a single representative voxel attribute \(\bar{a}_s\) via weighted summation:
    \small{$\bar{a}_s =\sum_{j \in \mathcal{G}_s} m_j \, a_j$}.
    The final memory bank is thus parameterized by the aggregated voxel attributes $\left\{ \left( \bar{\bm{\mu}}_s, \bar{o}_s, \bar{\bm{q}}_s, \bar{\bm{s}}_s, \bar{\bm{x}}_s \right) \right\}_{s=1}^S$. This strategy dramatically reduces the primitive count while preserving the gradient flow, ensuring that our incremental memory updates remain both geometrically accurate and computationally scalable. For notational simplicity, we omit the bar accent ($\bar{\cdot}$) in the remainder of the paper and refer to these aggregated voxel attributes simply as Gaussian parameters.

\section{Details of Static and Temporal Memory Retrieval}
    \label{sec:mem_ret}
    \textbf{Static Memory Retrieval.}
    For static background elements, we employ standard 3D Gaussian Splatting. Each Gaussian is parameterized by opacity $\alpha$, covariance $\boldsymbol{\Sigma}$, center $\bm{\mu}$, and the latent feature $\bm{x}$. The covariance is decomposed into scaling $\bm{S}$ and rotation $\bm{R}$ as $\boldsymbol{\Sigma} = \bm{R} \bm{S} \bm{S}^\top \bm{R}^\top$.
    Given the viewing transformation $\bm{W}_{i+1}$ and Jacobian $\bm{J}$, the covariance is transformed to $\boldsymbol{\Sigma}' = \bm{J} \bm{W}_{i+1} \boldsymbol{\Sigma} \bm{W}_{i+1}^\top \bm{J}^\top$. The blending weight for static elements is determined by the 2D projected spatial density:
    \small{
    \begin{equation}
        \label{eq:static_weight}
        w_j = G(\bm{p}_j) \alpha_j,
    \end{equation}
    }
    where \small{$G(\bm{p}) = \exp\left(-\frac{1}{2}(\bm{p} - \bm{\mu}')^\top {\boldsymbol{\Sigma}'}^{-1} (\bm{p} - \bm{\mu}')\right)$} and $\bm{\mu}'$ is the projected 2D mean.
    
    \textbf{Temporal Memory Retrieval.}
    To recall dynamic content consistent with the scene's evolution, we employ 4D Gaussian Splatting, predicting the object state at timestamp $t_{i+1}$. We extend the primitives with temporal quaternion $\bm{q}_{j,t}$ and scale $s_{j,t}$ to form 4D Gaussians centered at $\bm{\mu} = (\mu_x, \mu_y, \mu_z, \mu_t)$. The 4D covariance utilizes a diagonal scaling $\bm{S}$ and a 4D rotation matrix $\bm{R}$ constructed from left/right unit quaternions $\bm{q}_l = [s_l, \bm{v}_l^\top]^\top$ and $\bm{q}_r = [s_r, \bm{v}_r^\top]^\top$:
    \small{
    \begin{equation}
        \bm{R} = 
        \begin{bmatrix}
            s_l & -\bm{v}_l^\top \\
            \bm{v}_l & s_l \mathbf{I} + [\bm{v}_l]_\times
        \end{bmatrix}
        \begin{bmatrix}
            s_r & -\bm{v}_r^\top \\
            \bm{v}_r & s_r \mathbf{I} - [\bm{v}_r]_\times
        \end{bmatrix},
    \end{equation}
    }
    where $s_{\{\cdot\}} \in \mathbb{R}$ and $\bm{v}_{\{\cdot\}} \in \mathbb{R}^3$ denote the scalar and vector parts of the quaternions, respectively. The notation $[\cdot]_\times$ represents the $3 \times 3$ skew-symmetric matrix operator equivalent to the cross product, and $\mathbf{I}$ is the identity matrix.
    The recall process follows the slicing approach~\cite{yang2023real}. The weight $w_j$ for dynamic elements incorporates both the conditional spatial distribution and the marginal temporal probability:
    \small{
    \begin{equation}
        \label{eq:dynamic_weight}
        w_j = G(\bm{x}_j | t_{i+1}) G(t_{i+1}) \alpha_j.
    \end{equation}
    }
    The marginal probability is $G(t) = \mathcal{N}(t; \mu_t, \boldsymbol{\Sigma}_{4,4})$, and the conditional spatial term is defined as $G(\bm{x} | t) = \exp\left(-\frac{1}{2} (\bm{x} - \bm{\mu}_{\bm{x}|t})^\top \boldsymbol{\Sigma}_{\bm{x}|t}^{-1} (\bm{x} - \bm{\mu}_{\bm{x}|t})\right)$. The conditional mean and covariance are derived from the multivariate Gaussian properties: $\bm{\mu}_{\bm{x}|t} = \bm{\mu}_{1:3} + \bm{\Sigma}_{1:3,4} \bm{\Sigma}_{4,4}^{-1} (t_{i+1} - \mu_t)$ and $\bm{\Sigma}_{\bm{x}|t} = \bm{\Sigma}_{1:3,1:3} - \bm{\Sigma}_{1:3,4} \bm{\Sigma}_{4,4}^{-1} \bm{\Sigma}_{4,1:3}$.

\section{Implementation Details}
    \label{sec:imple}
    Our \ourMethod{} is implemented in PyTorch, utilizing the CUDA-optimized \texttt{gsplat} library~\cite{ye2025gsplat} for 3D Gaussian Splatting (3DGS) rendering. For the Pixel-Aligned Latent Writing module in Latent Gaussian Memory (\secref{sec:latent_memory}), we employ a ViT-Large~\cite{dosovitskiy2020image} encoder and a ViT-Base decoder, equipped with DPT prediction heads. This module is initialized with pretrained weights from CUT3R~\cite{wang2025cut3r} to facilitate accurate initial point cloud estimation. Our training pipeline proceeds in two stages: Latent Gaussian Memory construction and autoregressive video diffusion. In the first stage, we freeze the backbone and exclusively optimize the Gaussian and global heads. We sample video subsequences ranging from 5 to 81 frames to predict explicit Gaussian representations, using subsequent frames for supervision. All input images are resized to $512 \times 288$. In the second stage, we build our \ourMethod{} upon the Wan2.1-I2V-14B-480P model~\cite{wan2025}. To ensure deployment flexibility, we employ Low-Rank Adaptation (LoRA) for fine-tuning; this design allows users to seamlessly inject our method into private or customized models. Detailed hyperparameters are provided in \tabref{tab:hyperparameters}. All experiments were conducted on a cluster of 8 NVIDIA H200 GPUs, each equipped with 144 GB of memory. The reconstruction training in the first stage and the diffusion training in the second stage each take approximately 31 hours under this setup. During inference, generating an 81-frame video requires about 14 minutes, which is comparable to other Wan2.1 models at the 14B scale.

\begin{table}[htbp]
    \centering
    \begin{threeparttable}
        \caption{\textbf{Hyperparameter settings for DiT training}. Parameters are categorized into optimization strategies, model architecture, and deviation learning configurations.}
        \label{tab:hyperparameters}
        \small 
        \begin{tabularx}{0.85\textwidth}{llX}
            \toprule
            \textbf{Parameter} & \textbf{Value} & \textbf{Description} \\
            \midrule
            \multicolumn{3}{l}{\textbf{\textit{Optimization \& Training Strategy}}} \\
            Learning rate & $2.0 \times 10^{-5}$ & Learning rate for Adam optimizer \\
            Max epochs & 20 & Maximum number of training epochs \\
            Gradient clipping & 1.00 & Maximum norm for gradient clipping \\
            Gradient accumulation & 1 & Number of steps for gradient accumulation \\
            Training strategy & deepspeed\_stage\_2 & Distributed training strategy (DeepSpeed) \\
            Data workers & 1 & Number of workers for data loading \\
            Gradient checkpointing & Yes & Activation checkpointing for memory efficiency \\
            Checkpointing offload & No & Offload checkpoint activations to CPU \\
            \midrule
            \multicolumn{3}{l}{\textbf{\textit{Model Architecture \& LoRA Configuration}}} \\
            Architecture & LoRA & Type of fine-tuning architecture \\
            LoRA rank ($r$) & 128 & Rank dimension for Low-Rank Adaptation \\
            LoRA alpha ($\alpha$) & 128 & Scaling factor for LoRA \\
            LoRA init & Kaiming & Initialization scheme for LoRA weights \\
            LoRA position & q, k, v, o, ffn.0, ffn.2 & Transformer modules applying LoRA \\
            Frame resolution & $480 \times 832$ & Spatial resolution ($H \times W$) of frames \\
            Video frames & 5--81 & Range of sequence length for training samples \\
            \midrule
            \multicolumn{3}{l}{\textbf{\textit{Deviation Learning \& Dynamic Deviation Archive}}} \\
            Warmup iterations & 20 & Warmup steps for deviation gathering \\
            Deviation $p$ & 0.9 & Probability of deviation injection \\
            Clean input $p$ & 0.5 & Probability of conditioning on clean history \\
            Timestamp grids & 50 & Number of denoising timestamp bins in Dynamic Deviation Archive \\
            Archive capacity $Z$ & 500 & Capacity of each archive cell in Dynamic Deviation Archive \\
            \bottomrule
        \end{tabularx}
    \end{threeparttable}
\end{table}

\section{Runtime, Memory, and Robustness Analysis}
    \label{sec:runtime_memory_robustness}

    \begin{table}[t]
        \centering
        \caption{\textbf{Runtime and memory costs.} We report costs for a standard 81-frame sequence at $512 \times 288$ resolution.}
        \label{tab:efficiency}
        \small
        \setlength{\tabcolsep}{5pt}
        \resizebox{\linewidth}{!}{%
        \begin{tabular}{lcccc}
            \toprule
            Method / Component & GPU Memory & CPU Memory & Runtime & Notes \\
            \midrule
            Standard 3D-aware memory & $\sim$500 MB & -- & $\sim$93 ms / frame & $>$11M Gaussians \\
            Latent Gaussian Memory & $\sim$100 MB & -- & 76.34 ms / frame & dual-level pruning \\
            Dynamic Deviation Archive & 0.6 MB / frame & $\sim$22 GB & 0.266 ms / step & sampled CPU-to-GPU transfer \\
            \bottomrule
        \end{tabular}
        }
    \end{table}

    \noindent\textbf{Runtime and memory costs.}
    \ourMethod{} is efficient in both GPU memory and runtime, introducing only negligible overhead beyond the base video diffusion model. The Dynamic Deviation Archive is stored on CPU and occupies about 22 GB of CPU memory. During training, we transfer only the sampled deviation from CPU to GPU, which costs about 0.6 MB for each frame. Each one-step deviation approximation takes about 0.266 ms, since it performs only one local approximation, reuses existing forward-pass activations, and is computed in a \texttt{no\_grad} context.
    
    For a standard 81-frame sequence at $512 \times 288$ resolution, the Latent Gaussian Memory uses only about 100 MB of GPU memory, with an average runtime of 76.34 ms per frame. Compared with prior 3D-aware memory methods such as VMem~\cite{li2025vmem} and WorldWarp~\cite{kong2025worldwarp}, this is substantially more efficient. A standard per-pixel 3D-aware memory implementation would otherwise produce over 11 million Gaussians and require roughly 500 MB of GPU memory. Our efficiency comes from the dual-level pruning strategy in \secref{sec:mem_compress}, which prunes the 3D representation from both the 2D and 3D perspectives. Overall, this reduces memory usage by about 81\% and runtime by about 18\%. The detailed cost breakdown is provided in \tabref{tab:efficiency}.

    \noindent\textbf{Cost of maintaining the Dynamic Deviation Archive.}
    Maintaining and updating the Dynamic Deviation Archive introduces very small overhead in practice. In our implementation, archive maintenance costs at most about 350 ms per training iteration, while one full training iteration takes about 11.1 s. This corresponds to only about 3.2\% overhead in the worst case. Importantly, when training the current frame, we do not run an additional forward pass over historical frames. We simply sample stored deviations from the archive and inject them into historical latents before memory construction, as described in \secref{sec:deviation_archive}. Since the archive is implemented as a dictionary indexed by autoregressive stage and denoising timestamp, sampling and lookup are lightweight and their cost is negligible compared with the diffusion training step.

    \noindent\textbf{Efficiency and role of memory recall.}
    Our memory recall mechanism directly renders the stored Gaussians into the target view through Gaussian Splatting (\secref{sec:mem_ret} and \eqref{eq:latent_memory_recall}). This operation is efficient in practice: real-time 3DGS systems and optimized splatting libraries commonly report rendering speeds above 100 FPS~\cite{kerbl20233dgs,ye2025gsplat}, so the recall step adds only modest overhead relative to the 14B diffusion backbone. Although adjacent frames are often visually similar, small viewpoint changes can still reveal newly visible or previously unobserved regions, and those regions may have been better captured from earlier frames. The recall module retrieves such historical information as a view-aligned condition, helping the generated frame remain consistent with the accumulated history. This is consistent with the broader use of memory in autoregressive video generation and world modeling, including Self-Forcing~\cite{huang2025self-forcing}, WorldMem~\cite{xiao2025worldmem}, WorldWarp~\cite{kong2025worldwarp}, and VMem~\cite{li2025vmem}.

    \begin{table}[t]
        \centering
        \caption{\textbf{Scaling to longer videos.} We report total inference time using our 14B model on one NVIDIA H200 GPU.}
        \label{tab:long_video_runtime}
        \small
        \setlength{\tabcolsep}{8pt}
        \begin{tabular}{lcc}
            \toprule
            Sequence length & Total time & Average time / frame \\
            \midrule
            81 frames & 14.0 min & 10.4 s \\
            301 frames & 55.3 min & 11.0 s \\
            \bottomrule
        \end{tabular}
    \end{table}

    \noindent\textbf{Scaling to longer videos.}
    The computational cost of \ourMethod{} scales linearly with the number of generated frames. Standard full-sequence video generation models often rely on bidirectional attention, where each frame attends to every other frame, leading to $O(T^2)$ temporal attention cost for a $T$-frame sequence. In contrast, our autoregressive Dreamer predicts one new frame at a time and conditions only on the anchor frame, the previous frame, the recalled memory context, and the current control, as formulated in \eqref{eq:overview_infer_condition}. It therefore avoids attention over all generated frames and has $O(T)$ rollout complexity. Memory writing and recall are also performed once per generated frame, with the representation size controlled by the pruning strategy in \secref{sec:mem_compress}. Empirically, generating an 81-frame video takes about 14 minutes, while generating a 301-frame video takes 55.3 minutes on one NVIDIA H200 GPU, as shown in \tabref{tab:long_video_runtime}. This near-linear scaling indicates that the memory mechanism does not introduce disproportionate overhead for longer videos.

    \noindent\textbf{Robustness to noisy geometry, occlusion, and dynamics.}
    The Dynamic Deviation Archive stores latent-space deviations synthesized by the current generator and reuses them to corrupt historical latents during training. These deviation-corrupted histories expose the Dreamer to imperfect memory states similar to those encountered at inference, improving robustness to drift, noisy geometry, occlusion, and dynamic disturbances. However, when reconstruction quality is severely degraded and recoverable geometry is no longer available, our method may still fail, similar to other 3D-aware world modeling approaches that fundamentally rely on usable geometry. We discuss this limitation further in \secref{sec:limitation}.

    \noindent\textbf{Non-rigid dynamics.}
    Scenes with significant non-rigid motion are challenging because dynamic deviation patterns can be difficult to distinguish from true scene dynamics. Our method uses 4DGS-based temporal memory (\secref{sec:mem_ret}) to model dynamic content, and prior 4DGS studies suggest that Gaussian-based space-time representations can capture non-rigid motion in practice~\cite{luiten2023dynamic,wu20234d,yang2023real}. Nevertheless, our 4DGS component does not fully solve extreme non-rigid dynamics. The primary focus of this paper is to address drift from latent--RGB cycling and training--inference mismatch, rather than to design a dedicated module for highly non-rigid phenomena such as fluids, fire, or topology-changing motion. Our current training and experiments mainly cover static scenes and dynamic but relatively structured motion. Stronger performance on highly non-rigid scenarios would likely require dedicated datasets with reliable pose annotations and explicit training on such cases, which remain scarce. We view systematic evaluation and modeling of such dynamics as important future work.

    \noindent\textbf{Scope of the training--inference gap analysis.}
    We do not intend to attribute all long-horizon failures solely to the training--inference gap. Model capacity, scene complexity, control difficulty, reconstruction quality, and occlusion can also affect long-horizon generation. Our analysis in \secref{sec:deviation_archive} focuses on accumulated drift as one important failure mode of autoregressive rollout. The fact that drift can still occur with a high-capacity 14B backbone suggests that scaling model capacity alone is insufficient. We emphasize the training--inference gap because it is a common issue in autoregressive video generation, as also discussed by Self-Forcing~\cite{huang2025self-forcing} and Diffusion Forcing~\cite{chen2025diffusion}. Deviation Learning specifically addresses this factor by exposing the model to error-corrupted memory states during training, thereby reducing accumulated drift at inference time. Improving model capacity and handling more complex scenarios remain important and complementary directions.

    \noindent\textbf{Faithfulness of the one-step approximation.}
    We quantified the one-step deviation approximation against ODE-based simulation and found that the synthesized deviations reach about 58.7\% of the error magnitude obtained by explicit ODE simulation. As discussed in \secref{sec:deviation_archive}, the most accurate way to obtain deviations is to simulate inference by solving the ODE. However, this is computationally prohibitive during training, because it requires running the full iterative sampling chain, e.g., 50 or 100 integration steps, for every historical frame in each training iteration. Faithfully matching multi-step diffusion rollout errors with only a one-step approximation is therefore inherently difficult.
    
    This motivates the Dynamic Deviation Archive: rather than relying on a single approximation to perfectly reproduce full rollout error, we continuously collect diverse deviations throughout training. The archive uses an $\ell_2$-distance-based update rule to remove near-duplicate deviations while preserving diverse error modes. As a result, the model is exposed to a broad spectrum of realistic drift patterns and learns to recover from unexpected deviations more robustly at inference time. The ablation results in \tabref{tab:ablation} (Row B vs. Row D) validate that this deviation learning design is effective in practice.

    \noindent\textbf{Large deviations and hallucinated directions.}
    When the input is heavily corrupted, the one-step predicted velocity may point in an unreliable or hallucinated direction. This is one reason we maintain the Dynamic Deviation Archive instead of relying on a single deterministic approximation. The archive is intended to collect a diverse set of realistic error patterns arising during generation, including mild deviations as well as more severe or hallucination-like failures. The one-step approximation therefore does not need to be exact for every corrupted input; it needs to provide diverse corrupted conditions for training. The training objective in \eqref{eq:overview_loss} then encourages the Dreamer to map deviation-corrupted conditions back to the clean ground-truth target, forcing the model to learn an internal correction mechanism. If the archive contained only simple or homogeneous deviations, the model would be exposed to a much narrower set of failure modes and would be more likely to drift under harder corrupted inputs at inference time. That said, if the corruption becomes extremely large or falls far outside the archive distribution, the current approximation may still fail. We regard this as a limitation and an important direction for future work.

    \begin{table}[t]
        \centering
        \caption{\textbf{Additional simple corruption baselines.} We replace our synthesized deviation with simple corruptions under the same setting.}
        \label{tab:simple_corruption}
        \small
        \setlength{\tabcolsep}{8pt}
        \begin{tabular}{lcccc}
            \toprule
            Method & PSNR $\uparrow$ & SSIM $\uparrow$ & LPIPS $\downarrow$ & FID $\downarrow$ \\
            \midrule
            Blur & 14.13 & 0.271 & 0.391 & 47.80 \\
            Color shift & 14.31 & 0.276 & 0.387 & 47.40 \\
            Ours & \textbf{16.89} & \textbf{0.651} & \textbf{0.351} & \textbf{16.82} \\
            \bottomrule
        \end{tabular}
    \end{table}

    \noindent\textbf{Simple corruption baselines.}
    Gaussian noise is a common corruption baseline, so our main ablation reports this variant in \tabref{tab:ablation}. Following this concern, we additionally test two simple corruption baselines, blur and color shift, by replacing our synthesized deviation with each corruption under the same setting. As shown in \tabref{tab:simple_corruption}, these simple augmentations remain substantially worse than our method. This indicates that the archive does more than inject generic corruption: it stores structured, model-induced deviation patterns that better match autoregressive rollout errors.

\section{More Related Work}
    \label{sec:more_related}
    
    \paragraph{Feed-forward Reconstruction.}
    Foundational models like DUSt3R and MASt3R~\cite{dust3r_cvpr24,leroy2024mast3r} have achieved promising results by directly estimating point clouds from image pairs in a single pass, effectively handling challenging low-texture regions. To scale this capability to video sequences, Fast3R~\cite{yang2025fast3r} extends processing to thousands of frames simultaneously, while VGGT~\cite{wang2025vggt} introduces a generalized framework capable of extracting multiple 3D attributes from variable input lengths. StreamVGGT~\cite{zhuo2025streaming} caches historical keys and values in a causal transformer framework to maintain a persistent representation over long horizons without exploding computational costs; however, its computational and memory usage still grow redundantly over time. Recurrent architectures like CUT3R~\cite{wang2025cut3r} maintain a constant-sized memory state to ensure low inference costs, but they often suffer from forgetting earlier frames, leading to performance degradation as the sequence length increases. By implementing a simple yet efficient modification to the CUT3R architecture, TTT3R~\cite{chen2025ttt3r} explores a closed-form state update rule that enhances length generalization, allowing the model to reason over thousands of views while keeping memory and computation costs consistently low.
    
    \paragraph{Gaussian Splatting.}
    Gaussian Splatting has recently emerged as a powerful representation for real-time neural rendering. Kerbl et al.\cite{kerbl20233dgs} pioneered 3D Gaussian Splatting (3DGS), which models scenes using collections of 3D Gaussians, achieving real-time high-resolution rendering with state-of-the-art visual quality. This breakthrough has sparked widespread interest and numerous extensions. Several works~\cite{wu20234d,luiten2023dynamic,yang2023real,duan20244drotorgs,duan20244d} have adapted 3DGS to dynamic scenes. Beyond novel-view synthesis, Gaussian-based and closely related 3D field representations have also been explored for surface reconstruction~\cite{chen2023neusg,Huang2DGS2024,Yu2024GOF,chen2024vcr} and semantic or open-vocabulary scene understanding~\cite{qin2024langsplat,chen2023gnesf,wang2024gov}. However, most existing methods rely on per-scene optimization and lack generalization. To overcome this limitation, several studies~\cite{pixelsplat,chen2024mvsplat,chen2024mvsplat360,gps,transplat,hisplat,freesplat} proposed feed-forward 3DGS frameworks that reconstruct scenes from sparse-view images by predicting pixel-aligned Gaussian parameters and unprojecting them into 3D space, supervised by the interpolated views. These approaches, however, depend on accurate camera poses and significant pose overlap. To mitigate this, NoPoseSplat~\cite{ye2024noposplat} and Splatt3R~\cite{smart2024splatt3r} integrate Dust3r-based~\cite{dust3r_cvpr24,mast3r_arxiv24} geometry estimation to eliminate pose dependency. 
    
\section{Limitation}
    \label{sec:limitation}
    Our method shares a common limitation with many 3D-aware world modeling approaches in that its performance depends on the quality of the underlying 3D reconstruction. When reliable geometry cannot be recovered—due to severe occlusions, reflective surfaces, rapid motion, or insufficient viewpoint diversity—the Latent Gaussian Memory may become inaccurate or incomplete, which in turn can degrade long-horizon generation quality. While our framework mitigates temporal drift through robust memory usage and deviation-aware training, it does not remove the fundamental dependency on successful 3D reconstruction. In addition, our framework is not fully end-to-end, as it separates memory construction and generative modeling into distinct components. This modular design improves interpretability and stability but may limit joint optimization across memory and generation. An interesting direction for future work is to explore more unified architectures, potentially leveraging diffusion models to jointly perform memory formation and long-horizon generation within a single end-to-end framework.

\section{Impact Statements}
    This work focuses on algorithmic advances in long-horizon world modeling and does not involve the collection of new data. All training and evaluation are conducted using publicly available datasets, without the use of any private, sensitive, or personally identifiable data. The proposed methods are primarily designed for scene-level and environment-centric modeling, rather than human-centered applications. As a result, we do not anticipate direct ethical or societal risks related to privacy, data ownership, or human subject impact. As with other generative video and world-modeling techniques, potential misuse could include generating misleading scene-level simulations or over-relying on simulated rollouts in downstream decision-making, so deployment should be paired with dataset provenance checks and application-specific validation. We believe the contributions of this work are largely technical and methodological, supporting future research in simulation and world modeling under responsible research practices.

\section{More Qualitative Results}
    \label{sec:more_qualitative}

\begin{figure*}[t]
    \centering
    \includegraphics[width=\textwidth]{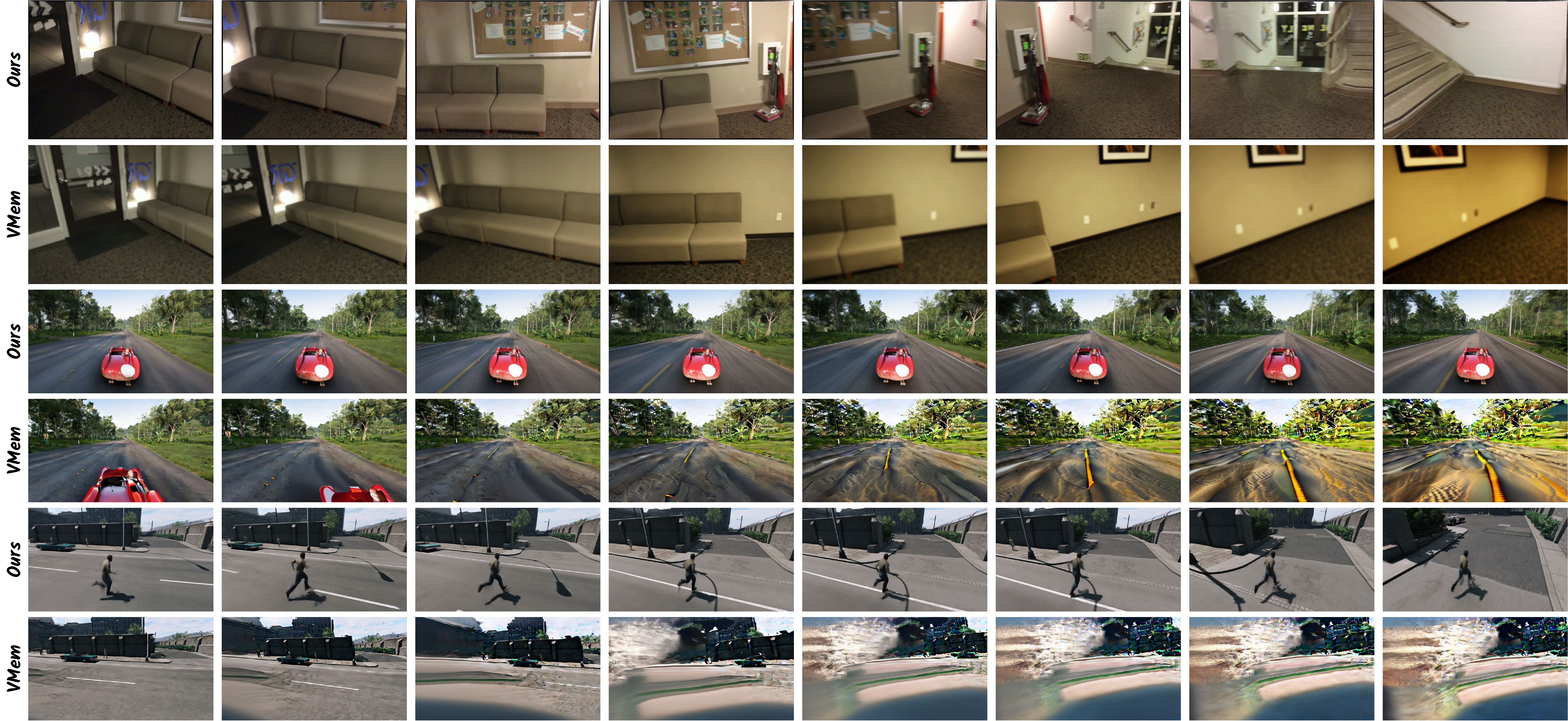}
    \captionof{figure}{
    \textbf{Qualitative results on static long scenes from ScanNet (300 frames) and dynamic scenes from OmniWorldGame (80 frames).} The top two rows show experiments on ScanNet, while the bottom four rows present comparisons on OmniWorldGame. Displayed frames are randomly sampled from the early, middle, and late stages of the sequences. Compared to the state-of-the-art baseline VMem, which also utilizes a 3D memory mechanism, our approach successfully avoids the color drift issue. This demonstrates the effectiveness of our proposed latent-memory inheritance and Deviation Learning. Additional comparisons will be included in future versions.
    }
    \vspace{-0.3cm}
\end{figure*}

    \begin{figure*}[t]
        \centering
        \includegraphics[width=\textwidth]{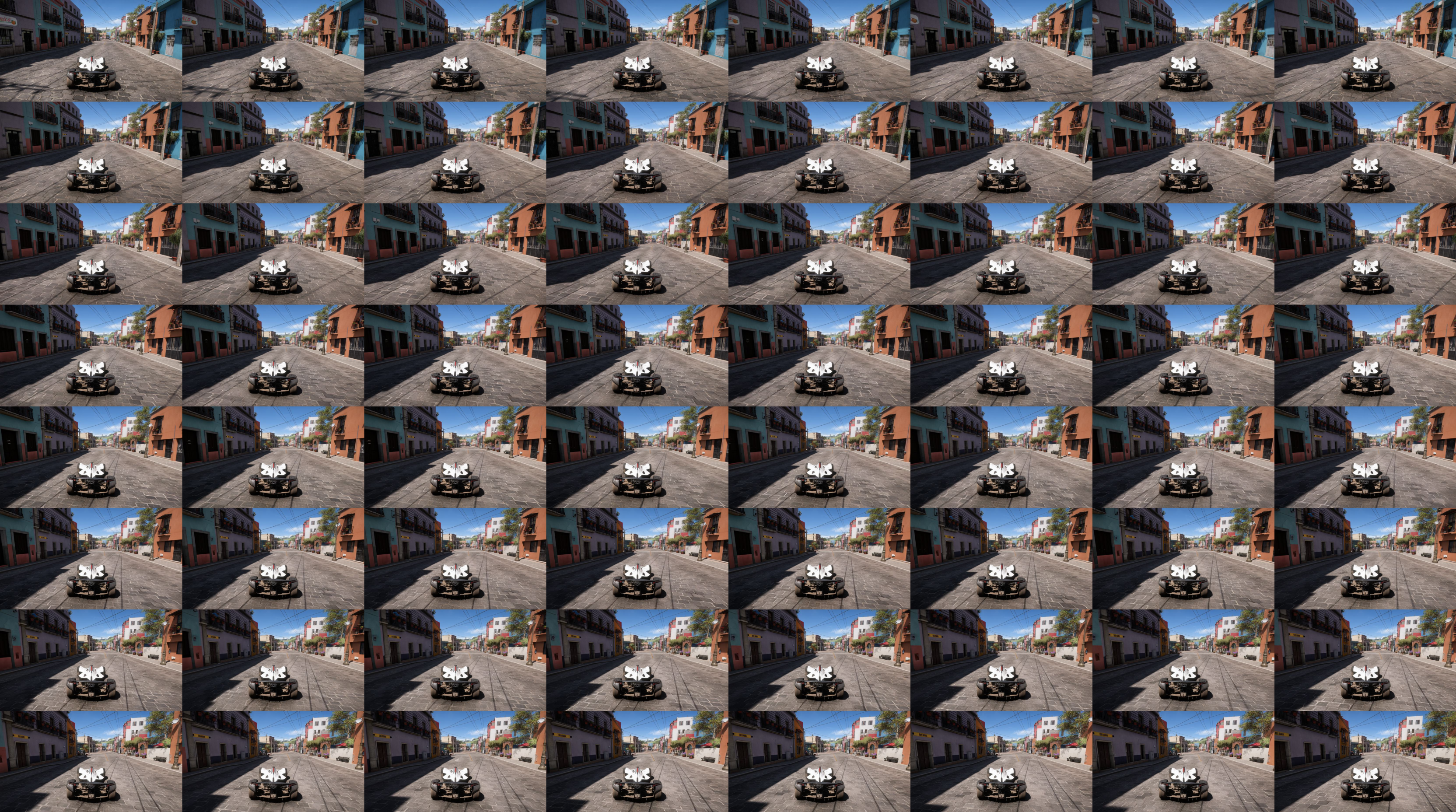}
        \captionof{figure}{
        \textbf{Qualitative results on OmniWorldGame.} We visualize 64 randomly sampled frames from an 80-frame \textbf{dynamic scene}.
        }
        \vspace{-0.3cm}
    \end{figure*}
    
    \begin{figure*}[t]
        \centering
        \includegraphics[width=\textwidth]{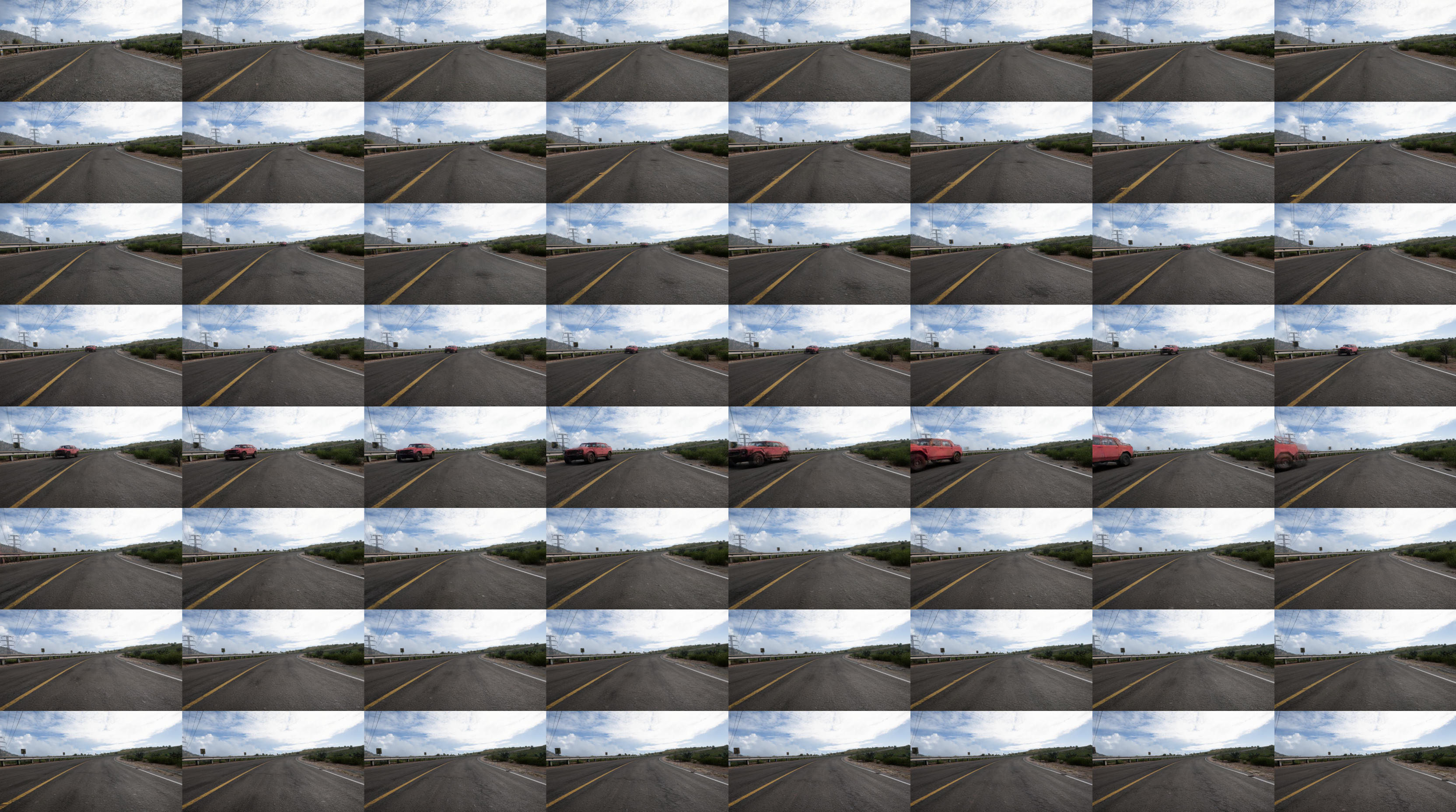}
        \captionof{figure}{
        \textbf{Qualitative results on OmniWorldGame.} We visualize 64 randomly sampled frames from an 80-frame \textbf{dynamic scene}.
        }
        \vspace{-0.3cm}
    \end{figure*}
    
    \begin{figure*}[t]
        \centering
        \includegraphics[width=\textwidth]{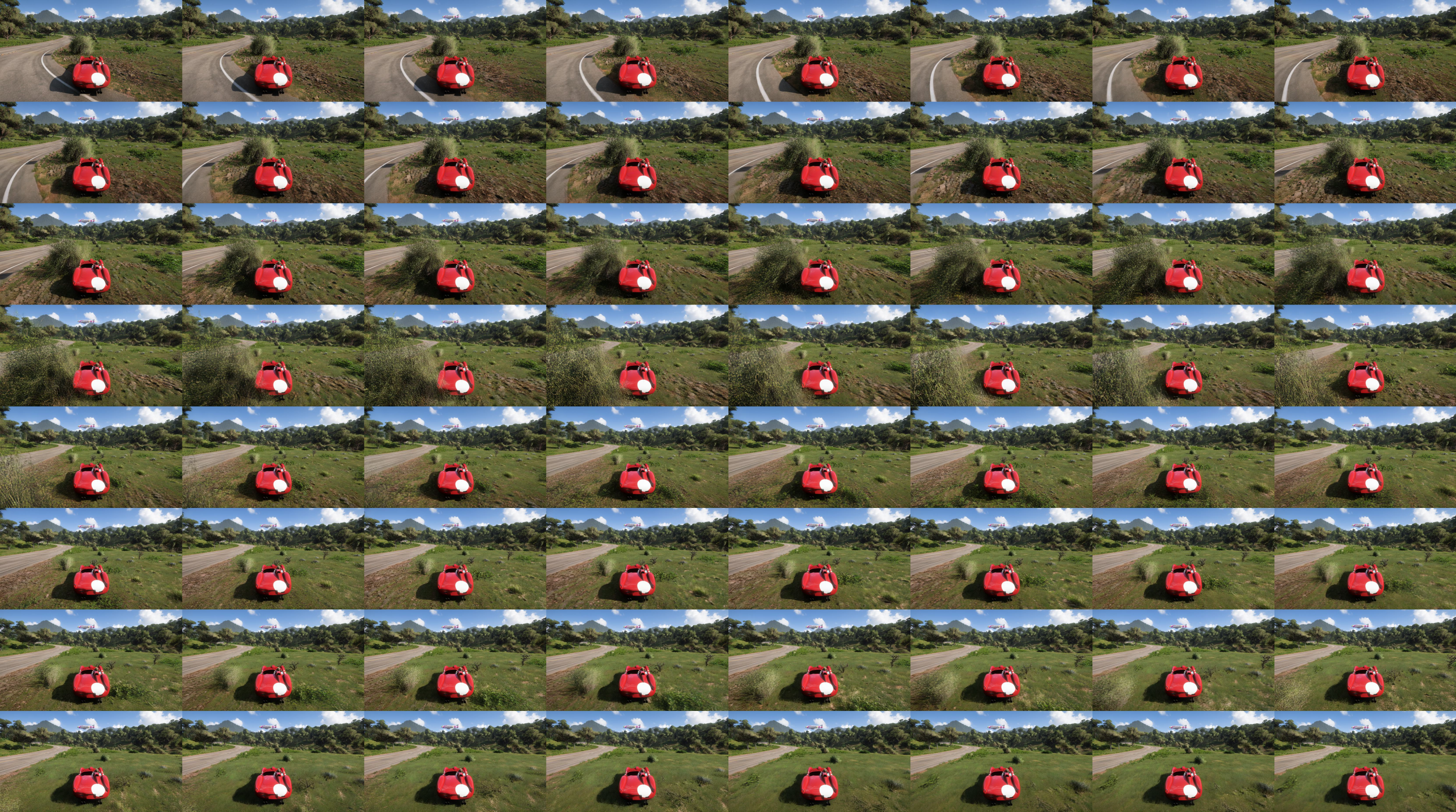}
        \captionof{figure}{
        \textbf{Qualitative results on OmniWorldGame.} We visualize 64 randomly sampled frames from an 80-frame \textbf{dynamic scene}.
        }
        \vspace{-0.5cm}
    \end{figure*}
    
    \begin{figure*}[t]
        \centering
        \includegraphics[width=\textwidth]{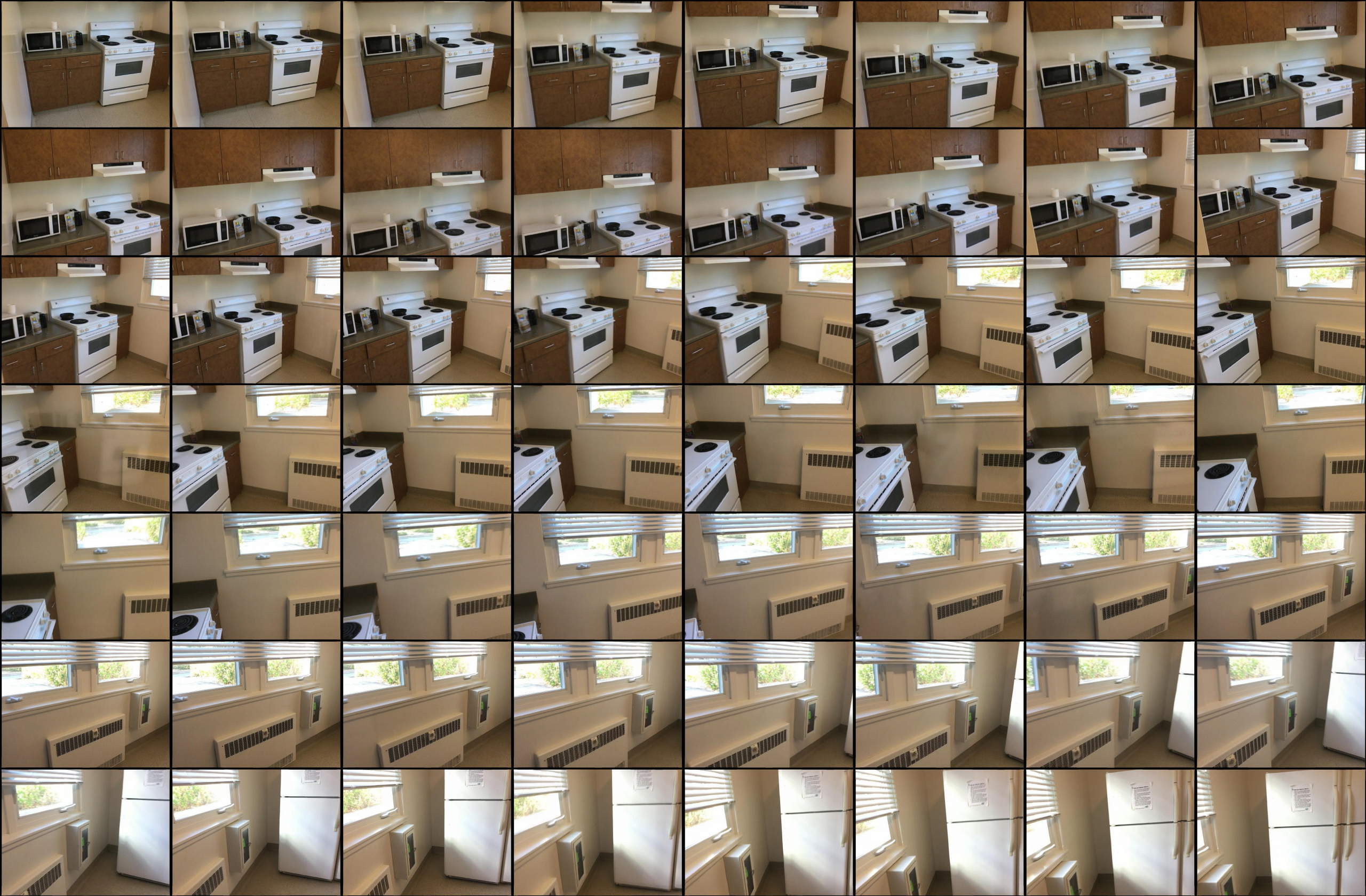}
        \captionof{figure}{
        \textbf{Qualitative results on ScanNet.} We visualize 56 frames uniformly sampled from a 300-frame sequence of a \textbf{long static scene}.
        }
        \vspace{-0.3cm}
    \end{figure*}

    \begin{figure*}[t]
        \centering
        \includegraphics[width=\textwidth]{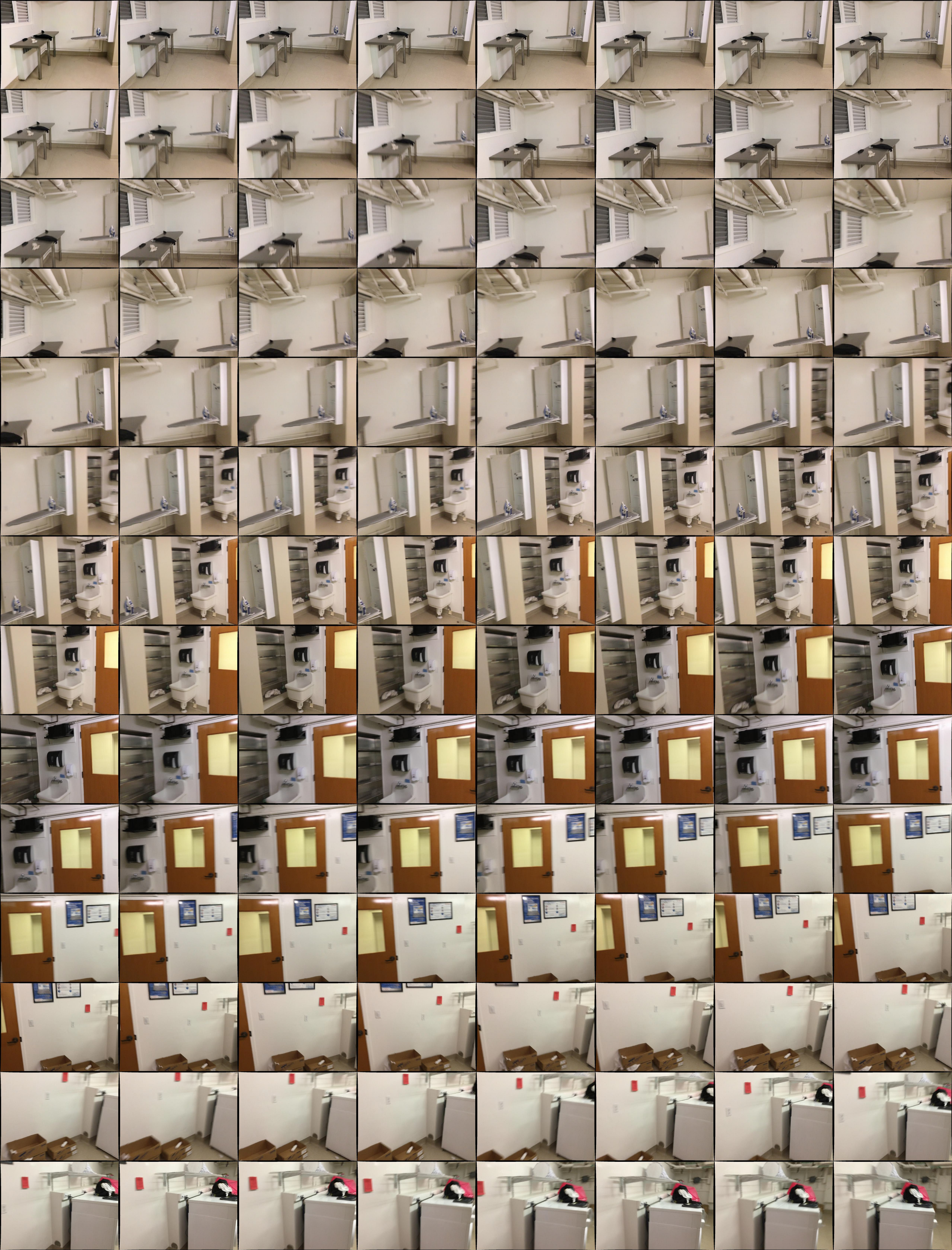}
        \captionof{figure}{
        \textbf{Qualitative results on ScanNet.} We visualize 112 frames uniformly sampled from a 300-frame sequence of a \textbf{long static scene}.
        }
    \end{figure*}

    \begin{figure*}[t]
        \centering
        \includegraphics[width=\textwidth]{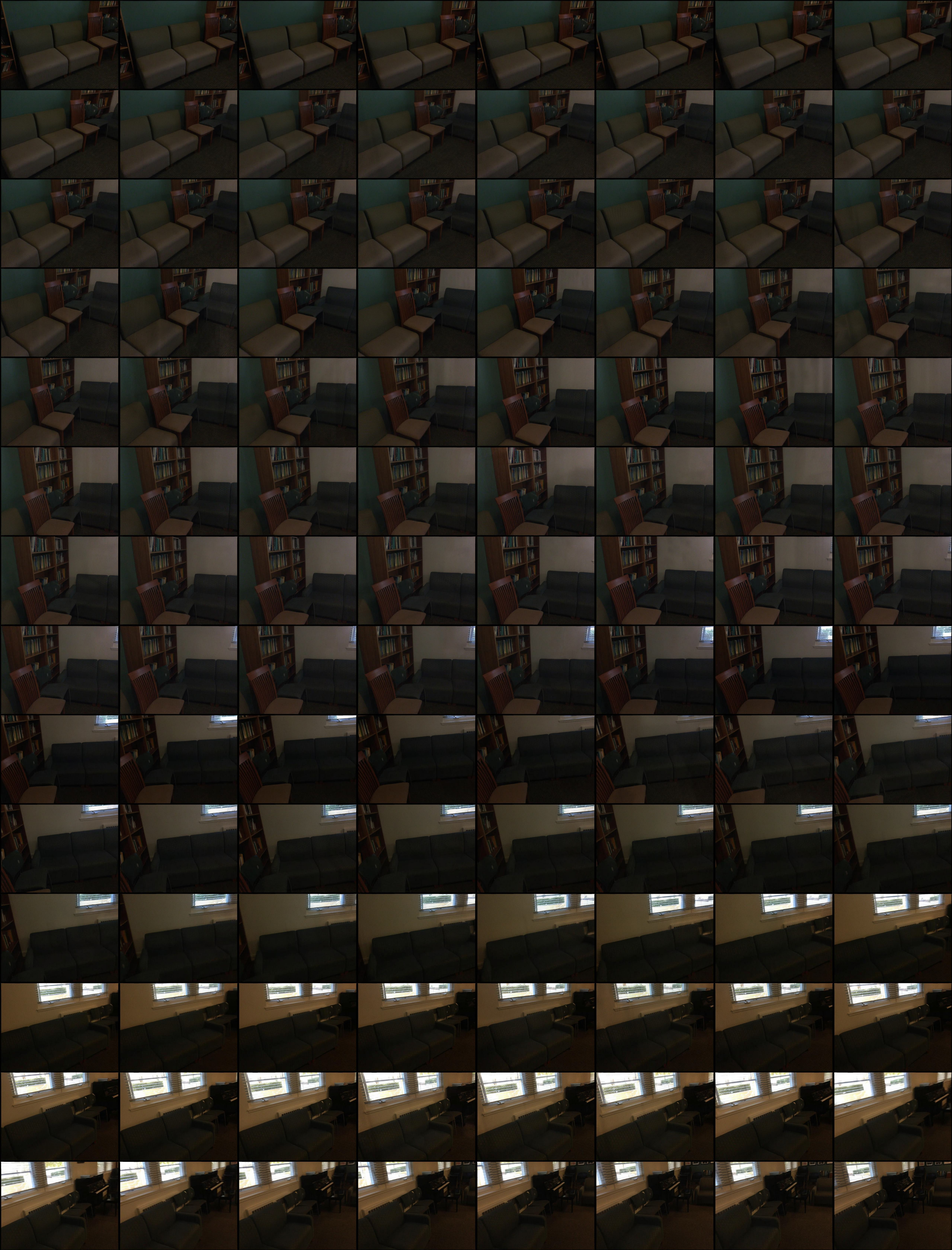}
        \captionof{figure}{
        \textbf{Qualitative results on ScanNet.} We visualize 112 frames uniformly sampled from a 300-frame sequence of a \textbf{long static scene}.
        }
        \vspace{-0.3cm}
    \end{figure*}
    
    \begin{figure*}[t]
        \centering
        \includegraphics[width=\textwidth]{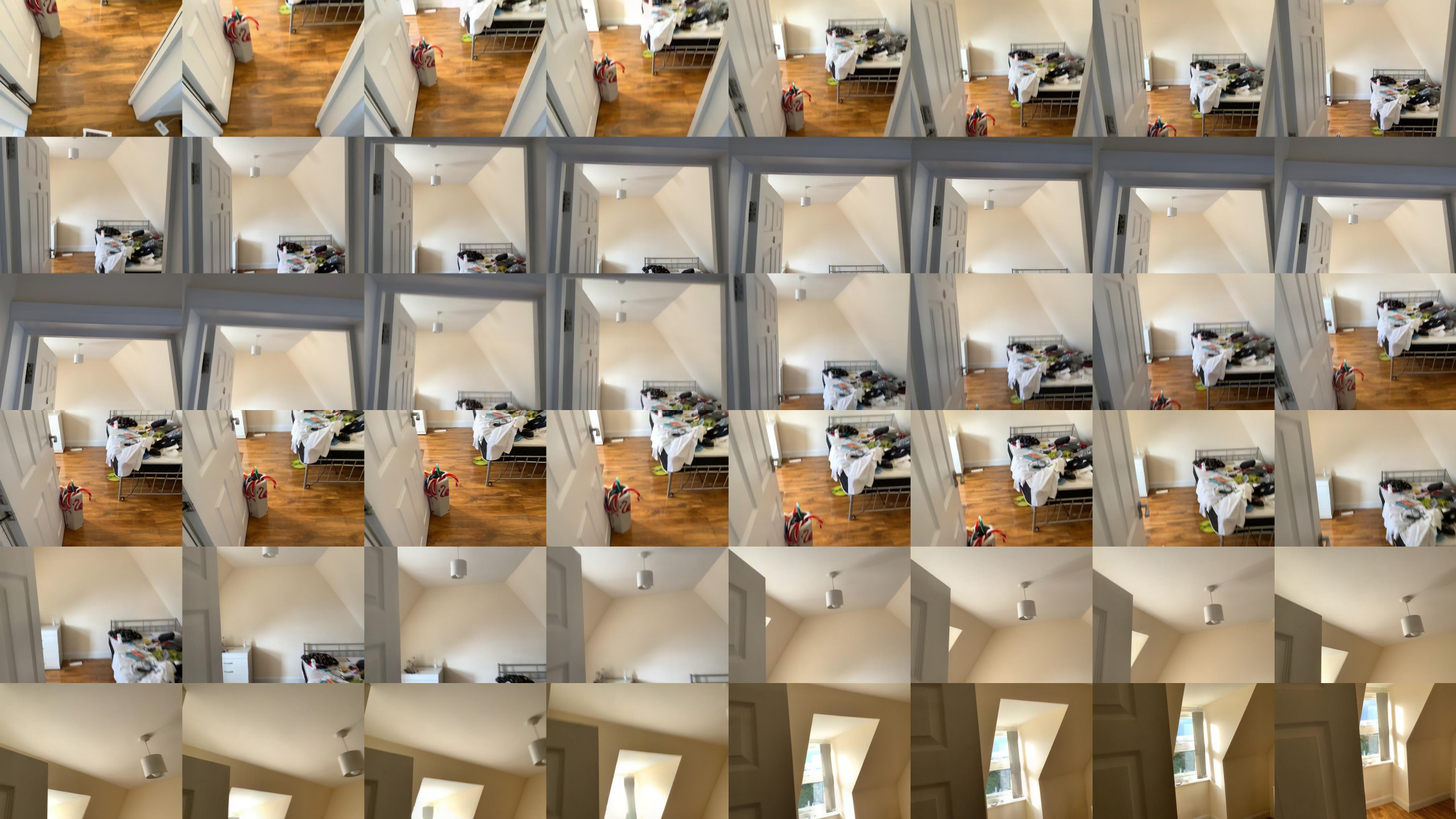}
        \captionof{figure}{
        \textbf{Qualitative results on an out-of-domain scene.} We visualize 48 uniformly sampled frames from an 80-frame sequence to demonstrate generalization.
        }
        \vspace{-0.3cm}
    \end{figure*}

    \begin{figure*}[t]
        \centering
        \includegraphics[width=\textwidth]{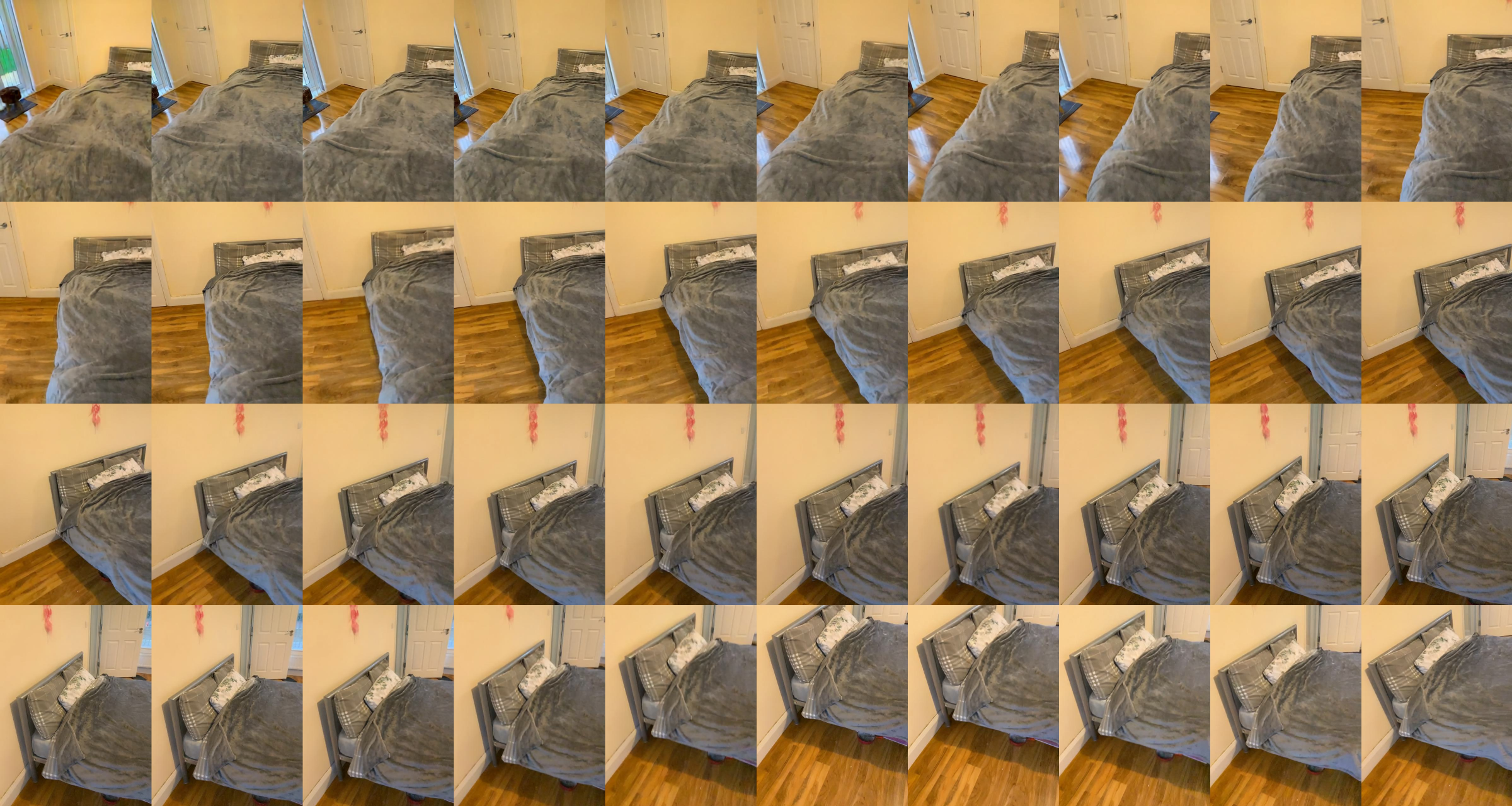}
        \captionof{figure}{
        \textbf{Qualitative results on an out-of-domain scene.} We visualize 40 uniformly sampled frames from an 80-frame sequence to demonstrate generalization.
        }
        \vspace{-0.3cm}
    \end{figure*}

    \begin{figure*}[t]
        \centering
        \includegraphics[width=\textwidth]{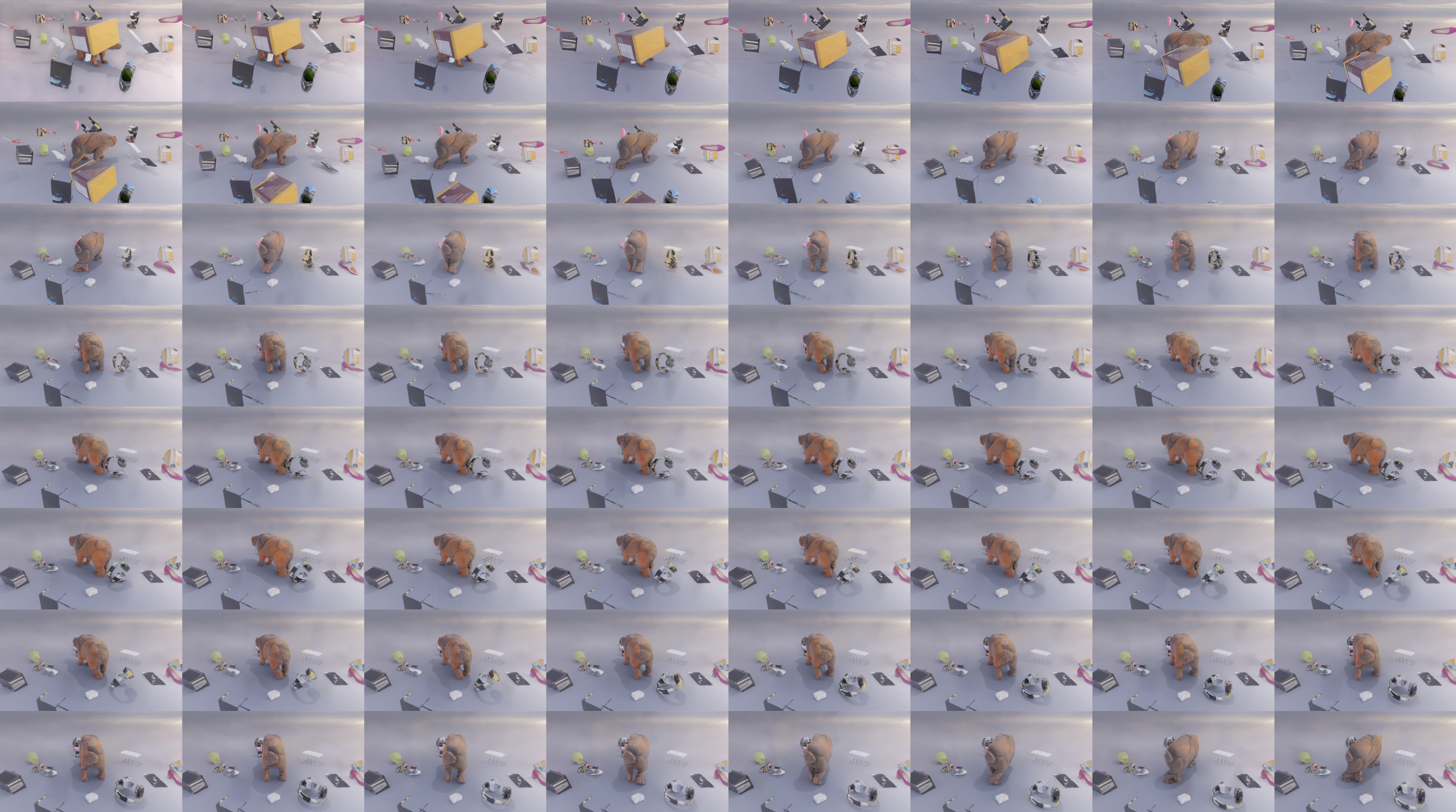}
        \captionof{figure}{
        \textbf{Qualitative results on an out-of-domain dynamic scene.} We visualize 64 uniformly sampled frames from an 80-frame sequence to demonstrate generalization.
        }
    \end{figure*}

    \clearpage
    \section*{NeurIPS Paper Checklist}

\begin{enumerate}

\item {\bf Claims}
    \item[] Question: Do the main claims made in the abstract and introduction accurately reflect the paper's contributions and scope?
    \item[] Answer: \answerYes{}
    \item[] Justification: The abstract and \secref{sec:intro} state the scope, assumptions, and contributions of Robust Dreamer. The method in \secref{sec:method} and experiments in \secref{sec:exp} support the claimed improvements in long-horizon action-controlled video generation.

\item {\bf Limitations}
    \item[] Question: Does the paper discuss the limitations of the work performed by the authors?
    \item[] Answer: \answerYes{}
    \item[] Justification: The appendix includes a dedicated Limitation section discussing dependence on 3D reconstruction quality and the modular, not fully end-to-end, design. These limitations clarify failure cases such as occlusion, reflective surfaces, rapid motion, and insufficient viewpoint diversity.

\item {\bf Theory assumptions and proofs}
    \item[] Question: For each theoretical result, does the paper provide the full set of assumptions and a complete (and correct) proof?
    \item[] Answer: \answerNA{}
    \item[] Justification: The paper does not present formal theoretical results or proofs. The equations in \secref{sec:method} define the model, memory, archive, and training objective rather than theorem statements.

    \item {\bf Experimental result reproducibility}
    \item[] Question: Does the paper fully disclose all the information needed to reproduce the main experimental results of the paper to the extent that it affects the main claims and/or conclusions of the paper (regardless of whether the code and data are provided or not)?
    \item[] Answer: \answerYes{}
    \item[] Justification: The paper specifies the method, training objective, datasets, metrics, baselines, implementation details, hyperparameters, and compute resources in \secref{sec:method}, \secref{sec:settings}, \secref{sec:imple}, and \tabref{tab:hyperparameters}. These details provide the information needed to reproduce the main experimental comparisons at the level described in the submission.

\item {\bf Open access to data and code}
    \item[] Question: Does the paper provide open access to the data and code, with sufficient instructions to faithfully reproduce the main experimental results, as described in supplemental material?
    \item[] Answer: \answerNo{}
    \item[] Justification: The experiments use publicly available datasets, but the current submission does not include an open code release, anonymized repository, or command-level reproduction instructions. We plan to provide release instructions in a future version where possible.

\item {\bf Experimental setting/details}
    \item[] Question: Does the paper specify all the training and test details (e.g., data splits, hyperparameters, how they were chosen, type of optimizer) necessary to understand the results?
    \item[] Answer: \answerYes{}
    \item[] Justification: \secref{sec:settings} describes datasets, metrics, evaluation settings, and baselines, while \secref{sec:imple} and \tabref{tab:hyperparameters} provide model, optimization, training, resolution, sequence-length, and deviation-learning details.

\item {\bf Experiment statistical significance}
    \item[] Question: Does the paper report error bars suitably and correctly defined or other appropriate information about the statistical significance of the experiments?
    \item[] Answer: \answerNo{}
    \item[] Justification: The paper reports aggregate quantitative metrics in \tabref{tab:scannet}, \tabref{tab:dl3dv}, and \tabref{tab:ablation}, but does not include error bars, confidence intervals, or statistical significance tests. Repeated runs are computationally expensive for the 14B-scale video model, so this version focuses on fixed-protocol comparisons.

\item {\bf Experiments compute resources}
    \item[] Question: For each experiment, does the paper provide sufficient information on the computer resources (type of compute workers, memory, time of execution) needed to reproduce the experiments?
    \item[] Answer: \answerYes{}
    \item[] Justification: \secref{sec:imple} reports the use of 8 NVIDIA H200 GPUs with 144 GB memory each, the approximate 31-hour runtime for each training stage, and the inference time for generating an 81-frame video.
    
\item {\bf Code of ethics}
    \item[] Question: Does the research conducted in the paper conform, in every respect, with the NeurIPS Code of Ethics \url{https://neurips.cc/public/EthicsGuidelines}?
    \item[] Answer: \answerYes{}
    \item[] Justification: The research uses publicly available datasets and does not collect private, sensitive, or personally identifiable data, as discussed in the Impact Statements. The work is methodological and environment-centric rather than a human-subject study.

\item {\bf Broader impacts}
    \item[] Question: Does the paper discuss both potential positive societal impacts and negative societal impacts of the work performed?
    \item[] Answer: \answerYes{}
    \item[] Justification: The Impact Statements discuss the positive role of the method in responsible world-modeling research and explain why direct privacy, data ownership, and human-subject risks are limited. They also discuss possible misuse through misleading scene-level simulations or over-reliance on simulated rollouts in downstream decision-making.
    
\item {\bf Safeguards}
    \item[] Question: Does the paper describe safeguards that have been put in place for responsible release of data or models that have a high risk for misuse (e.g., pre-trained language models, image generators, or scraped datasets)?
    \item[] Answer: \answerNA{}
    \item[] Justification: The paper does not introduce a scraped dataset or release a high-risk pretrained generative model in the current submission. The method is evaluated on existing public datasets, and no new high-risk asset release is described.

\item {\bf Licenses for existing assets}
    \item[] Question: Are the creators or original owners of assets (e.g., code, data, models), used in the paper, properly credited and are the license and terms of use explicitly mentioned and properly respected?
    \item[] Answer: \answerNo{}
    \item[] Justification: The paper cites the datasets, model backbones, and software components used in the work, including ScanNet, DL3DV, OmniWorldGame, Wan2.1, CUT3R-style reconstruction, and gsplat. The current manuscript does not yet explicitly list the licenses and terms of use for each asset.

\item {\bf New assets}
    \item[] Question: Are new assets introduced in the paper well documented and is the documentation provided alongside the assets?
    \item[] Answer: \answerNA{}
    \item[] Justification: The paper does not introduce or release a new dataset or other standalone research asset in the current submission. It proposes a method and evaluates it using existing datasets and model components.

\item {\bf Crowdsourcing and research with human subjects}
    \item[] Question: For crowdsourcing experiments and research with human subjects, does the paper include the full text of instructions given to participants and screenshots, if applicable, as well as details about compensation (if any)? 
    \item[] Answer: \answerNA{}
    \item[] Justification: The paper does not involve crowdsourcing or research with human subjects. It uses existing public datasets for training and evaluation.

\item {\bf Institutional review board (IRB) approvals or equivalent for research with human subjects}
    \item[] Question: Does the paper describe potential risks incurred by study participants, whether such risks were disclosed to the subjects, and whether Institutional Review Board (IRB) approvals (or an equivalent approval/review based on the requirements of your country or institution) were obtained?
    \item[] Answer: \answerNA{}
    \item[] Justification: The paper does not involve crowdsourcing or human-subject research, so IRB approval or equivalent review is not applicable.

\item {\bf Declaration of LLM usage}
    \item[] Question: Does the paper describe the usage of LLMs if it is an important, original, or non-standard component of the core methods in this research? Note that if the LLM is used only for writing, editing, or formatting purposes and does \emph{not} impact the core methodology, scientific rigor, or originality of the research, declaration is not required.
    \item[] Answer: \answerNA{}
    \item[] Justification: The core method does not use LLMs as an important, original, or non-standard component. The paper builds on video diffusion, latent Gaussian memory, and deviation-aware training rather than LLM-based methodology.

\end{enumerate}

\end{document}